\documentclass[10pt,twocolumn,letterpaper]{article}

\usepackage{iccv}
\usepackage{times}
\usepackage{epsfig}
\usepackage{graphicx}
\usepackage{amsmath}
\usepackage{gensymb}

\usepackage{xspace}
\usepackage{caption}
\usepackage{mathtools}
\usepackage{amsfonts}
\usepackage{booktabs}
\usepackage{graphicx}
\usepackage{soul}
\usepackage{color}
\usepackage{xspace}
\usepackage[subrefformat=parens]{subfig}

\usepackage{bbding}
\usepackage{pifont}
\usepackage{wasysym}
\usepackage{amssymb}
\usepackage{float}
\usepackage{dblfloatfix}
\usepackage{footnote}
\makesavenoteenv{tabular}
\makesavenoteenv{table}
\makesavenoteenv{figure}
\usepackage{authblk}

\usepackage{url}
\usepackage{placeins}
\usepackage[pagebackref=true,breaklinks=true,letterpaper=true,colorlinks,bookmarks=false]{hyperref}

 \iccvfinalcopy 

\def\iccvPaperID{3438} 

\newcommand{\xmark}{\ding{55}}

\ificcvfinal\fi

\def\OURS{DeeperCluster\xspace}
\def\etal{\emph{et al}\onedot}

\begin{document}

\title{Unsupervised Pre-Training of Image Features on Non-Curated Data}

\author[1,2]{Mathilde Caron}
\author[1]{Piotr Bojanowski}
\author[2]{Julien Mairal}
\author[1]{Armand Joulin}
\affil[1]{Facebook AI Research}
\affil[2]{Univ. Grenoble Alpes, Inria, CNRS, Grenoble INP, LJK, 38000 Grenoble, France}

\maketitle

\begin{abstract}
Pre-training general-purpose visual features with convolutional neural networks
without relying on annotations is a challenging and important task. Most
recent efforts in unsupervised feature learning have focused on either small or
highly curated datasets like ImageNet, whereas using non-curated raw datasets
was found to decrease the feature quality when evaluated on a transfer task.
Our goal is to bridge the performance gap between unsupervised methods trained on curated
data, which are costly to obtain, and massive raw datasets that are easily
available.
To that effect, we propose a new unsupervised approach which leverages self-supervision and clustering to capture complementary statistics from large-scale data.
We validate our approach on $96$ million
images from YFCC100M~\cite{thomee2015yfcc100m}, achieving state-of-the-art
results among unsupervised methods on standard benchmarks, which confirms
the potential of unsupervised learning when only non-curated raw data are available.
We also show that pre-training a supervised VGG-16 with our method achieves $74.9\%$ top-$1$ classification accuracy on the validation set of ImageNet, which is an improvement of $+0.8\%$ over the same network trained from scratch.
Our code is available at \url{https://github.com/facebookresearch/DeeperCluster}.
\end{abstract}

\section{Introduction}
Pre-trained convolutional neural networks, or convnets, are important components of image recognition applications~\cite{carreira2016human,chen2016deeplab,ren2015faster,weinzaepfel2013deepflow}.
They improve the generalization of models trained on a limited amount of data~\cite{sharif2014cnn} and speed up the training on applications when annotated data is abundant~\cite{he2018rethinking}.
Convnets produce good generic representations when they are pre-trained on large supervised datasets like ImageNet~\cite{deng2009imagenet}.
However, designing such fully-annotated datasets has required a significant effort from the research community in terms of data cleansing and manual labeling.
Scaling up the annotation process to datasets that are orders of magnitude bigger raises important difficulties.
Using raw metadata as an alternative has been shown to perform comparatively well~\cite{joulin2016learning,sun2017revisiting}, even surpassing ImageNet pre-training when trained on billions of images~\cite{mahajan2018exploring}.
However, metadata are not always available, and when they are, they do not necessarily cover the full extent of a dataset.
These difficulties motivate the design of methods that learn transferable features without using any annotation.

\begin{figure}[t!]
  \centering
  \includegraphics[width=0.495\linewidth]{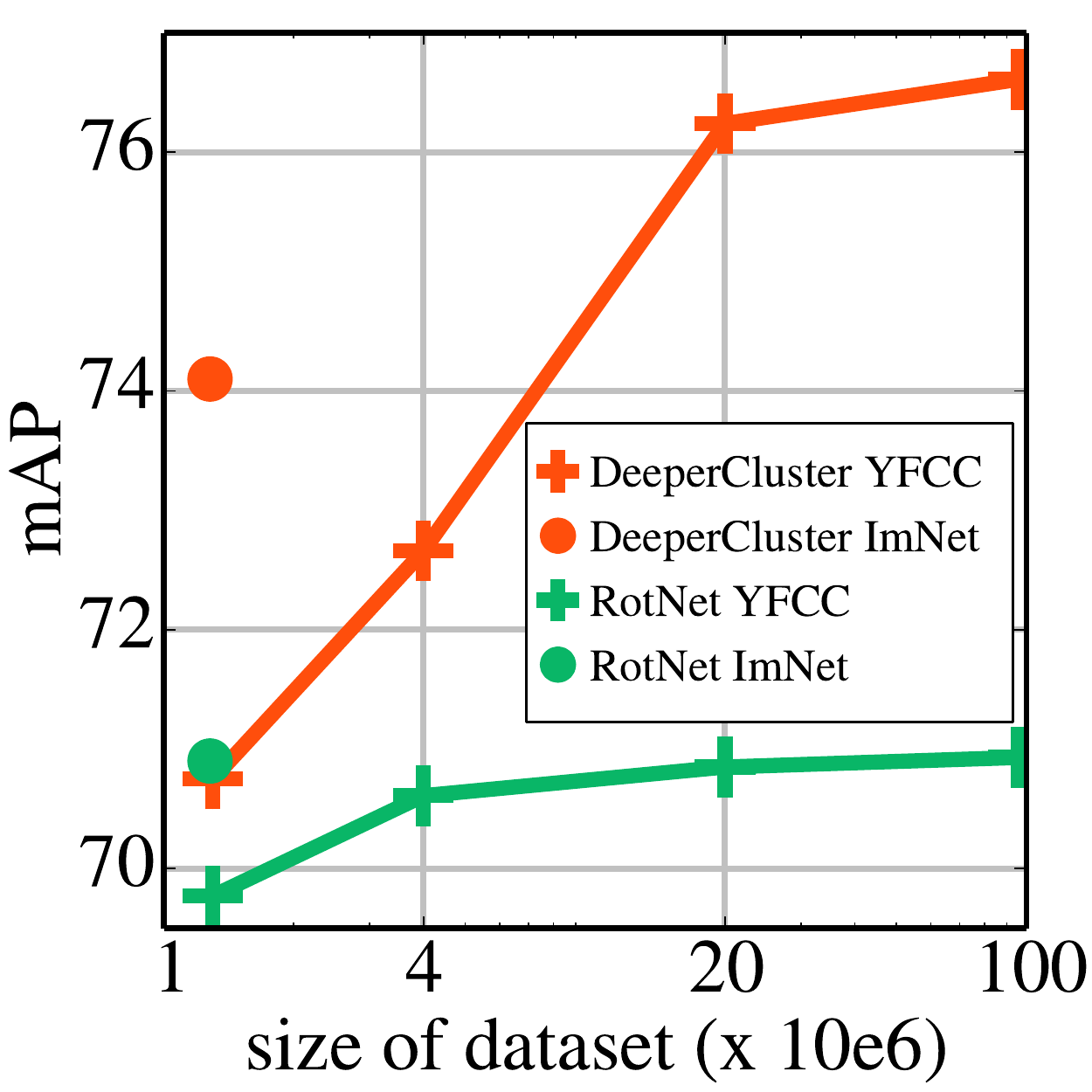}
  \includegraphics[width=0.495\linewidth]{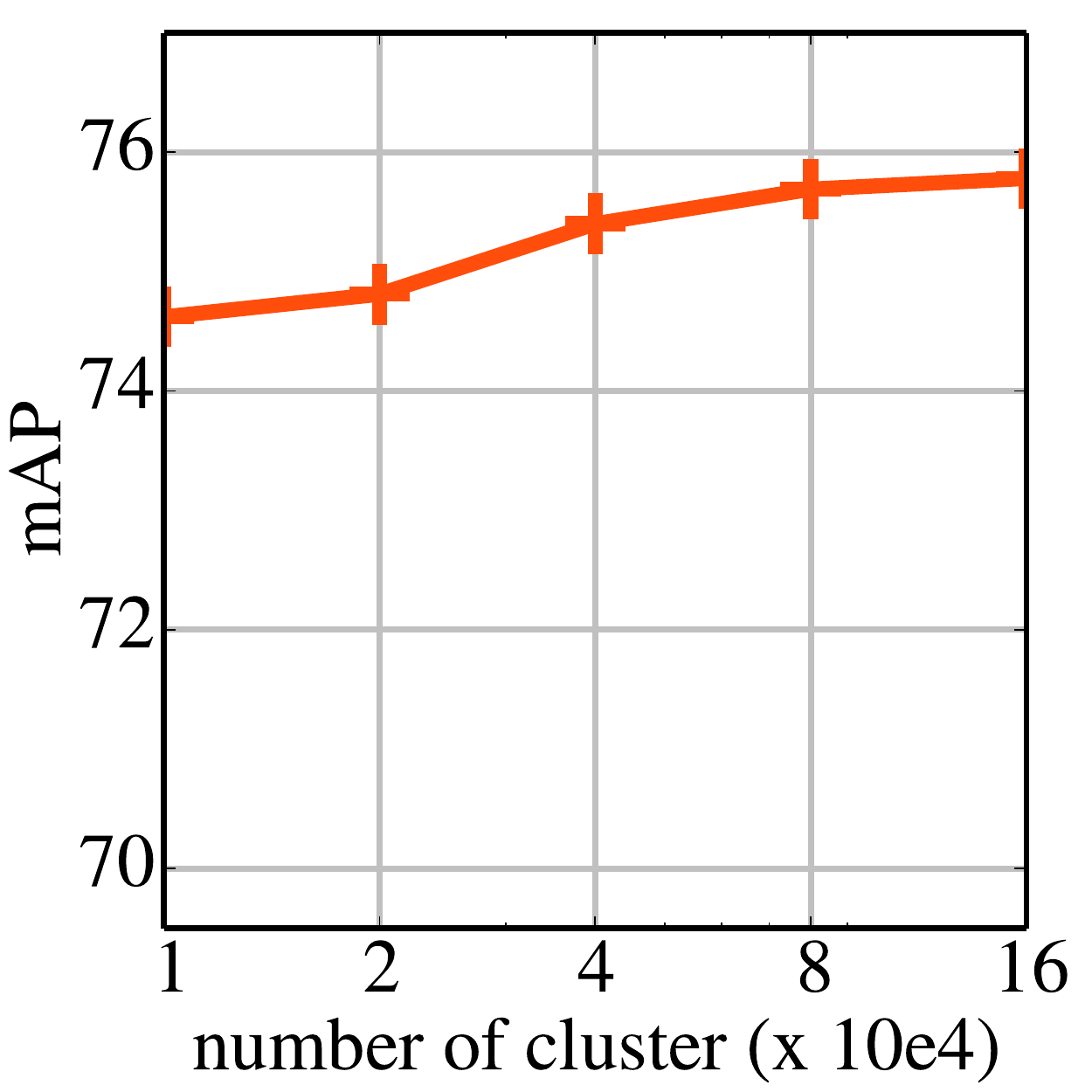}
  \caption{
  Influence of amount of data (\textit{left}) and number of clusters (\textit{right}) on the features quality.
  We report validation mAP on Pascal VOC classification task (\textsc{fc68} setting).
  }
  \label{fig:k}
\end{figure}

Recent works describing unsupervised approaches have reported performances that are closing the gap with their supervised counterparts~\cite{caron2018deep,gidaris2018unsupervised,zhang2019aet}.
However, the best performing unsupervised methods are trained on ImageNet, a curated dataset made of carefully selected images to form well-balanced and diversified classes~\cite{deng2009imagenet}.
Simply discarding the labels does not undo this careful selection, as it only removes part of the human supervision.
Because of that, previous works that have experimented with non-curated raw data report a degradation of the quality of features~\cite{caron2018deep,doersch2015unsupervised}.
In this work, we aim at learning good visual representations from unlabeled and non-curated datasets.
We focus on the YFCC100M dataset~\cite{thomee2015yfcc100m}, which contains $99$ million images from the Flickr photo-sharing website.
This dataset is unbalanced, with a ``long-tail'' distribution of hashtags contrasting with the well-behaved label distribution of ImageNet (see Appendix).
For example, \textit{guenon} and \textit{baseball} correspond to labels with $1300$ associated images in ImageNet, while there are respectively $226$ and $256,758$ images associated with these hashtags in YFCC100M.
Our goal is to understand if trading manually-curated data for scale leads to an improvement in the feature quality.

We propose a new unsupervised approach specifically designed to leverage large amount of raw data.
Indeed, training on large-scale non-curated data requires (i) model complexity to increase with dataset size; (ii) model stability to data distribution changes.
A simple yet effective solution is to combine methods from two domains of unsupervised learning: clustering and self-supervision.
Since clustering methods, like DeepCluster~\cite{caron2018deep}, build supervision from \textit{inter-image} similarities, the task at hand becomes inherently more complex when the number of images increases.
In addition, DeepCluster captures finer relations between images when the number of clusters scales with the dataset size.
Clustering approaches infer target labels at the same time as features are learned.
Thus, target labels evolve during training, making clustering-based approaches unstable.
Furthermore, these methods are sensitive to data distribution as they rely directly on cluster structure in the underlying data.
Explicitly dealing with unbalanced category distribution might be a solution but it assumes that we know the distribution of the latent classes.
We design our method without this assumption.
On the other hand, self-supervised learning~\cite{de1994learning} consists in designing a pretext task by predicting pseudo-labels automatically extracted from input signals~\cite{doersch2015unsupervised}.
In other words, self-supervised approaches, like RotNet~\cite{gidaris2018unsupervised}, leverage \textit{intra-image} statistics to build supervision, which are often independent of the data distribution.
However, the dataset size has little impact on the nature of the task and on the performance of the resulting features (see Figure~\ref{fig:k}).
A solution to leveraging larger datasets require manually increasing the difficulty of the self-supervision task~\cite{goyal2019scaling}.
Our approach automatically increases complexity through the clustering strategy.

\begin{table}[h]
  \centering
  \setlength{\tabcolsep}{1pt}
  \begin{tabular}{lcccccc}
    \toprule
	  && \footnotesize intra-image && \footnotesize inter-images && \footnotesize stable to distri- \\
	  \small Method && \footnotesize statistics && \footnotesize statistics. && \footnotesize bution change \\

    \midrule
	  \small Self-Sup (RotNet) && \checkmark && \xmark && \checkmark \\
    \midrule
	  \small Deep Clustering && \checkmark && \checkmark && \xmark \\
    \bottomrule
  \end{tabular}
\caption{Training on non-curated large-scale data requires model complexity to increase with dataset size and model stability to data distribution changes.
A simple solution is to combine self-supervision and clustering.}
\label{comp}
\end{table}

The novelty of our method lies in the combination of these two paradigms (Table~\ref{comp}) so that they benefit from one another.
Our approach, DeeperCluster, automatically generates targets by clustering the features of the entire dataset, under constraints derived from self-supervision.
Due to the ``long-tail'' distribution of raw non-curated data, processing huge datasets and learning a large number of targets is necessary, making the problem challenging from a computational point of view.
For this reason, we propose a hierachical formulation that is suitable for distributed training.
This enables the discovery of latent categories present in the ``tail'' of the image distribution.
While our framework is general, in practice we focus on combining the large rotation classification task of Gidaris~\etal~\cite{gidaris2018unsupervised} with the clustering approach of Caron~\etal~\cite{caron2018deep}.
Figure~\ref{fig:k} left shows that as we increase the number of training images, the quality of features improves to the point where it surpasses those trained without labels on curated datasets.
More importantly, we evaluate the quality of our approach as a pre-training step for ImageNet classification.
Pre-training a supervised VGG-16 with our unsupervised approach leads to a top-$1$ accuracy of $74.9\%$, which is an improvement of $+0.8\%$ over a model trained from scratch.
This shows the potential of unsupervised pre-training on large non-curated datasets as a way to improve the quality of visual features.


\section{Related Work}
\paragraph{Self-supervision.}
Self-supervised learning builds a pretext task from the input signal to train a model without annotation~\cite{de1994learning}.
Many pretext tasks have been proposed~\cite{jenni2018self,mahendran2018cross,wang2017transitive,wu2018unsupervised}, exploiting, amongst others,
spatial context~\cite{doersch2015unsupervised,kim2018learning,noroozi2016unsupervised,noroozi2018boosting,pathak2016context},
cross-channel prediction~\cite{larsson2016learning,larsson2017colorization,zhang2016colorful,zhang2017split},
or the temporal structure of videos~\cite{agrawal2015learning,pathak2017learning,wang2015unsupervised}.
Some pretext tasks explicitly encourage the representations to be either invariant or discriminative to particular types of input tranformations.
For example, Dosovitskiy~\etal~\cite{dosovitskiy2016discriminative} consider each image and its transformations as a class to enforce invariance to data transformations.
In this paper, we build upon the work of Gidaris~\etal~\cite{gidaris2018unsupervised} where the model encourages features to be discriminative for large rotations.
Recently, Kolesnikov~\etal~\cite{kolesnikov2019revisiting} have conducted an extensive benchmark of self-supervised learning methods on different convnet architectures.
As opposed to our work, they use curated datasets for pre-training.

\paragraph{Deep clustering.}
Clustering, along with density estimation and dimensionality reduction, is a family of standard unsupervised learning methods.
Various attempts have been made to train convnets using clustering~\cite{bautista2016cliquecnn,bojanowski2017unsupervised,caron2018deep,liao2016learning,wang2017unsupervised,xie2016unsupervised,yang2016joint}.
Our paper builds upon the work of Caron~\etal~\cite{caron2018deep}, in which $k$-means is used to cluster the visual representations.
Unlike our work, they mainly focus on training their approach using ImageNet without labels.
Recently, Noroozi~\etal~\cite{noroozi2018boosting} show that clustering can also be used as a form of distillation to improve the performance of networks trained with self-supervision. 
As opposed to our work, they use clustering only as a post-processing step and does not leverage the complementarity between clustering and self-supervision to further improve the quality of features.

\paragraph{Learning on non-curated datasets.}
Some methods~\cite{chen2015webly,gomez2017self,ni2015large} aim at learning visual features from non-curated data streams.
They typically use metadata such as hashtags~\cite{joulin2016learning,sun2017revisiting} or geolocalization~\cite{weyand2016planet} as a source of noisy supervision.
In particular, Mahajan~\etal~\cite{mahajan2018exploring} train a network to classify billions of Instagram images into predefined and clean sets of hashtags.
They show that with little human effort, it is possible to learn features that transfer well to ImageNet, even achieving state-of-the-art performance if finetuned.
As opposed to our work, they use an extrinsic source of supervision that had to be cleaned beforehand.


\section{Preliminaries}
In this work, we refer to the vector obtained at the penultimate layer of the convnet as a \emph{feature} or \emph{representation}.
We denote by $f_\theta$ the feature-extracting function, parametrized by a set of parameters $\theta$.
Given a set of images, our goal is then to learn a ``good'' mapping $f_{\theta^*}$.
By ``good'', we mean a function that produces general-purpose visual features that are useful on downstream tasks.

\subsection{Self-supervision}
In self-supervised learning, a pretext task is used to extract target labels directly from data~\cite{doersch2015unsupervised}.
These targets can take a variety of forms.
They can be categorical labels associated with a multiclass problem, as when predicting the transformation of an image~\cite{gidaris2018unsupervised,zhang2019aet} or the ordering of a set of patches~\cite{noroozi2016unsupervised}.
Or they can be continuous variables associated with a regression problem, as when predicting image color~\cite{zhang2016colorful} or surrounding patches~\cite{pathak2016context}.
In this work, we are interested in the former.
We suppose that we are given a set of $N$ images $\{x_1, \dots, x_N\}$ and we assign a pseudo-label $y_n$ in $\mathcal{Y}$ to each input $x_n$.
Given these pseudo-labels, we learn the parameters $\theta$ of the convet jointly with a linear classifier $V$ to predict pseudo-labels by solving the problem
\begin{equation}
  \label{eq:selfsup}
  \min_{\theta, V} \frac{1}{N} \sum_{n=1}^N \ell( y_n, V f_\theta(x_n)),
\end{equation}
where $\ell$ is a loss function.
The pseudo-labels $y_n$ are fixed during the optimization and the quality of the learned features entirely depends on their relevance.

\paragraph{Rotation as self-supervision.}
Gidaris~\etal~\cite{gidaris2018unsupervised} have recently shown that good features can be obtained when training a convnet to discriminate between different image rotations.
In this work, we focus on their pretext task, \textit{RotNet}, since its performance on standard evaluation benchmarks is among the best in self-supervised learning.
This pretext task corresponds to a multiclass classification problem with four categories: rotations in $\{0\degree, 90\degree, 180\degree, 270\degree\}$.
Each input $x_n$ in Eq.~(\ref{eq:selfsup}) is randomly rotated and associated with a target~$y_n$ that represents the angle of the applied rotation.

\subsection{Deep clustering}
Clustering-based approaches for deep networks typically build target classes by clustering visual features produced by convnets.
As a consequence, the targets are updated during training along with the representations and are potentially different at each epoch.
In this context, we define a latent pseudo-label $z_n$ in $\mathcal{Z}$ for each image $n$ as well as a corresponding linear classifier $W$.
These clustering-based methods alternate between learning the parameters $\theta$ and $W$ and updating the pseudo-labels $z_n$.
Between two reassignments, the pseudo-labels $z_n$ are fixed, and the parameters and classifier are optimized by solving
\begin{equation}
  \label{eq:mstep}
  \min_{\theta, W} \frac{1}{N} \sum_{n=1}^N \ell( z_n, W f_\theta(x_n)),
\end{equation}
which is of the same form as Eq.~(\ref{eq:selfsup}).
Then, the pseudo-labels $z_n$ can be reassigned by minimizing an auxiliary loss function.
This loss sometimes coincides with Eq.~(\ref{eq:mstep})~\cite{bojanowski2017unsupervised,xie2016unsupervised} but some works proposed to use another objective~\cite{caron2018deep,yang2016joint}.

\paragraph{Updating the targets with $k$-means.}
In this work, we focus on the framework of Caron~\etal~\cite{caron2018deep}, \textit{DeepCluster}, where latent targets are obtained by clustering the activations with $k$-means.
More precisely, the targets $z_n$ are updated by solving the following optimization problem:
\begin{equation}\label{eq:clustering}
  \min_{C\in\mathbb{R}^{d\times k}}  \sum_{n=1}^N \left[\min_{z_n\in \{0,1\}^k    ~\text{s.t.}~ z_n^\top {\mathbf 1} = 1} \| C z_n - f_\theta(x_n)\|_2^2 \right],
\end{equation}
$C$ is the matrix where each column corresponds to a centroid, $k$ is the number of centroids, and $z_n$ is a binary vector with a single non-zero entry.
This approach assumes that the number of clusters $k$ is known \emph{a priori}; in practice, we set it by validation on a downstream task (see Sec.~\ref{sec:exp-k}).
The latent targets are updated every $T$ epochs of stochastic gradient descent steps when minimizing the objective~(\ref{eq:mstep}).

Note that this alternate optimization scheme is prone to trivial solutions and controlling the way optimization procedures of both objectives interact is crucial.
Re-assigning empty clusters and performing a batch-sampling based on an uniform distribution over the cluster assignments are workarounds to avoid trivial parametrization~\cite{caron2018deep}.


\section{Method}
\label{sec:method}

In this section, we describe how we combine self-supervised learning with deep clustering in order to scale up to large numbers of images and targets.

\subsection{Combining self-supervision and clustering}

We assume that the inputs $x_1,\dots,x_N$ are rotated images, each associated with a target label $y_n$ encoding its rotation angle and a cluster assignment~$z_n$.
The cluster assignment changes during training along with the visual representations.
We denote by $\mathcal{Y}$ the set of possible rotation angles and by $\mathcal{Z}$, the set of possible cluster assignments.
A way of combining self-supervision with deep clustering is to add the losses defined in Eq.~(\ref{eq:selfsup}) and Eq.~(\ref{eq:mstep}).
However, summing these losses implicitly assumes that classifying rotations and cluster memberships are two independent tasks, which may limit the signal that can be captured.
Instead, we work with the Cartesian product space $\mathcal{Y} \times \mathcal{Z}$, which can potentially capture richer interactions between the two tasks.
We get the following optimization problem:
\begin{equation}
  \label{eq:naive}
  \min_{\theta, W} \frac{1}{N} \sum_{n=1}^N \ell(  y_n \otimes z_n, W f_\theta(x_n)).
\end{equation}
Note that any clustering or self-supervised approach with a multiclass objective can be combined with this formulation.
For example, we could use a self-supervision task that captures information about tiles permutations~\cite{noroozi2016unsupervised} or frame ordering in a video~\cite{wang2015unsupervised}.
However, this formulation does not scale in the number of combined targets, i.e., its complexity is $O(|\mathcal{Y}||\mathcal{Z}|)$.
This limits the use of a large number of cluster or a self-supervised task with a large output space~\cite{zhang2019aet}.
In particular, if we want to capture information contained in the tail of the distribution of non-curated dataset, we may need a large number of clusters.
We thus propose an approximation of our formulation based on a scalable hierarchical loss that it is designed to suit distributed training.

\subsection{Scaling up to large number of targets}

Hierarchical losses are commonly used in language modeling where the goal is to predict a word out of a large vocabulary~\cite{brown1992class}.
Instead of making one decision over the full vocabulary, these approaches split the process in a hierarchy of decisions, each with a smaller output space.
For example, the vocabulary can be split into clusters of semantically similar words, and the hierarchical process would first select a cluster and then a word within this cluster.

Following this line of work, we partition the target labels into a $2$-level hierarchy where we first predict a super-class and then a sub-class among its associated target labels.
The first level is a partition of the images into $S$ super-classes and we denote by $y_n$ the super-class assignment vector in $\{0,1\}^S$ of the image $n$ and by $y_{ns}$ the $s$-th entry of $y_n$.
This super-class assignment is made with a linear classifier~$V$ on top of the features.
The second-level of the hierarchy is obtained by partitioning \emph{within each super-class}.
We denote by $z^s_n$ the vector in $\{0,1\}^{k_s}$ of the assignment into~$k_s$ sub-classes for an image $n$ belonging to super-class $s$.
There are $S$ sub-class classifiers $W_1,\dots,W_S$, each predicting the sub-class memberships within a super-class $s$.
The parameters of the linear classifiers~$(V, W_1, \dots, W_S)$ and $\theta$ are jointly learned by minimizing the following loss function:
\begin{equation}\label{eq:sup}
\hspace{-5pt}\frac{1}{N} \sum_{n=1}^N \left[\ell\big(V f_\theta(x_n), y_n\big) {+}\sum_{s=1}^S y_{ns} \ell\left(W_s f_\theta(x_n) , z^s_n\right)\right],
\end{equation}
where $\ell$ is the negative log-softmax function. Note that an image that does not belong to the super-class $s$ does not belong either to any of its $k_s$ sub-classes.

\begin{figure}[t]
  \centering
  \includegraphics[width=\linewidth]{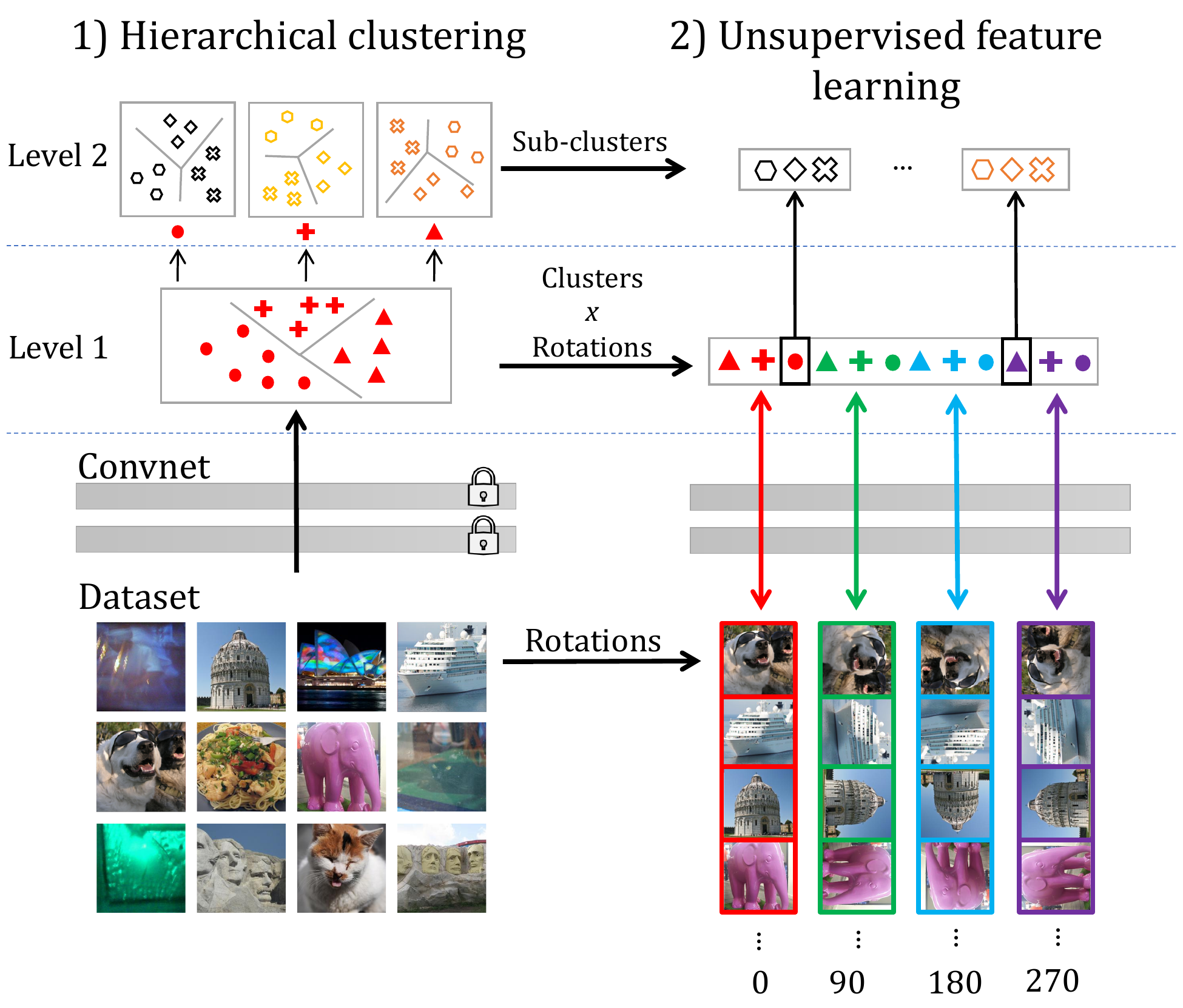}
  \caption{DeeperCluster alternates between a hierachical clustering of the features and learning the parameters of a convnet by predicting both the rotation angle and the cluster assignments in a single hierachical loss.}
   \vspace{-1em}
  \label{fig:method}
\end{figure}

\paragraph{Choice of super-classes.}
A natural partition would be to define the super-classes based on the target labels from the self-supervised task and the sub-classes as the labels produced by clustering.
However, this would mean that each image of the entire dataset would be present in each super-class (with a different rotation), which does not take advantage of the hierarchical structure to use a bigger number of clusters.

Instead, we split the dataset into $m$ sets by running $k$-means with $m$ centroids on the full dataset every $T$ epochs.
We then use the Cartesian product between the assignment to these $m$ clusters and the angle rotation classes to form the super-classes.
There are $4m$ super-classes, each associated with the subset of data belonging to the corresponding cluster ($N/m$ images if the clustering is perfectly balanced).
These subsets are then further split with $k$-means into $k$ subclasses.
This is equivalent to running a hierarchical $k$-means with rotation constraints on the full datasets to form our hierarchical loss.
We typically use $m=4$ and $k=80$k, leading to a total of $320$k different clusters split in $4$ subsets.
Our approach, ``\OURS'', shares similarities with DeepCluster but is designed to scale to larger datasets.
We alternate between clustering the non-rotated images features and training the network to predict both the rotation applied to the input data and its cluster assignment amongst the clusters corresponding to this rotation (Figure~\ref{fig:method}).

\paragraph{Distributed training.}
Building the super-classes based on data splits lends itself to a distributed implementation that scales well in the number of images.
Specifically, when optimizing Eq.~(\ref{eq:sup}), we form as many distributed communication groups of $p$ GPUs as the number of super-classes, i.e., $G=4m$.
Different communication groups share the parameters $\theta$ and the super-class classifier $V$, while the parameters of the sub-class classifiers $W_1,\dots,W_S$ are only shared within a communication group.
Each communication group $s$ deals only with the subset of images and the rotation angle associated with the super-class $s$.

\paragraph{Distributed $k$-means.}
Every $T$ epochs, we recompute the super and sub-class assignments by running two consecutive $k$-means on the entire dataset.
This is achieved by first randomly splitting the dataset across different GPUs.
Each GPU is in charge of computing cluster assignments for its partition, whereas
centroids are updated across GPUs.
We reduce communication between GPUs by sharing only the number of assigned elements for each cluster and the sum of their features.
The new centroids are then computed from these statistics.
We observe empirically that $k$-means converges in $10$ iterations.
We cluster $96$M features of dimension $4096$ into $m=4$ clusters using $64$ GPUs ($1$ minute per iteration).
Then, we split this pool of GPUs into $4$ groups of $16$ GPUs.
Each group clusters around $23$M features into $80$k clusters ($4$ minutes per iteration).

\subsection{Implementation details}

The loss in Eq.~(\ref{eq:sup}) is minimized with mini-batch stochastic gradient descent~\cite{bottou2012stochastic}.
Each mini-batch contains $3072$ instances distributed accross $64$ GPUs, leading to $48$ instances per GPU per minibatch~\cite{goyal2017accurate}.
We use dropout, weight decay, momentum and a constant learning rate of $0.1$.
We reassign clusters every $3$ epochs.
We use the Pascal VOC $2007$ classification task without finetuning as a downstream task to select hyper-parameters.
In order to speed up experimentations, we initialize the network with RotNet trained on YFCC100M.
Before clustering, we perform a whitening of the activations and $\ell_2$-normalize each of them.
We use standard data augmentations, i.e., cropping of random sizes and aspect ratios and horizontal flips~\cite{krizhevsky2012imagenet}).
We use the VGG-$16$ architecture~\cite{simonyan2014very} with batch normalization layers.
Following~\cite{bojanowski2017unsupervised,caron2018deep,paulin2015local}, we pre-process images with a Sobel filtering.
We train our models on the $96$M images from YFCC100M~\cite{thomee2015yfcc100m} that we managed to download.
We use this publicly available dataset for research purposes only.

\section{Experiments}
In this section we evaluate the quality of the features learned with \OURS on a variety of downstream tasks, such as classification or object detection.
We also provide insights about the impact of the number of images and clusters on the performance of our model.

\subsection{Evaluating unsupervised features}
We evaluate the quality of the features extracted from a convnet trained with \OURS on YFCC100M by considering several standard transfer learning tasks, namely image classification, object detection and scene classification.

\begin{table}[t!]
  \centering
  \setlength{\tabcolsep}{1pt}
  \begin{tabular}{@{}lc@{}c@{}c@{\hspace{0.2em}}c@{}c@{}c@{}c@{}}
    \toprule
	  & &\phantom{ee}& \multicolumn{2}{c}{Classif.} &\phantom{ee}& \multicolumn{2}{c}{Detect.} \\
                      \cmidrule{4-5} \cmidrule{7-8}
    Method & Data && \textsc{fc68} & \textsc{all} && \textsc{fc68} & \textsc{all} \\
    \midrule
    ImageNet labels                                 & INet        && $89.3^{\phantom{\dagger}}$ & $89.2^{\phantom{\dagger}}$  && $66.3^{\phantom{\dagger}}$  & $70.3^{\phantom{\dagger}}$ \\
	  Random                                          & --          && $10.1^{\phantom{\dagger}}$  & $49.6^{\phantom{\dagger}}$  && $\phantom{0}5.4^{\phantom{\dagger}}$  & $55.6^{\phantom{\dagger}}$ \\
    \midrule
	  \multicolumn{6}{l}{\textit{Unsupervised on curated data}}
    \vspace{0.3em} \\
	  Larsson~\etal~\cite{larsson2017colorization}    & INet+Pl.      && -- & $77.2^\dagger$    && $49.2^{\phantom{\dagger}}$ & $59.7^{\phantom{\dagger}}$ \\
	  Wu~\etal~\cite{wu2018unsupervised}              & INet          && -- & --       && -- & $60.5^\dagger$ \\
	  Doersh~\etal~\cite{doersch2015unsupervised}     & INet          && $54.6^{\phantom{\dagger}}$   & $78.5^{\phantom{\dagger}}$         && $38.0^{{\phantom{\dagger}}}$ & $62.7^{{\phantom{\dagger}}}$ \\
	  Caron~\etal~\cite{caron2018deep}                & INet          && $78.5^{\phantom{\dagger}}$ & $82.5^{\phantom{\dagger}}$ && $58.7^{\phantom{\dagger}}$  & $65.9^\dagger$ \\
    \midrule
	  \multicolumn{6}{l}{\textit{Unsupervised on non-curated data}}
    \vspace{0.3em} \\
	  Mahendran~\etal~\cite{mahendran2018cross}       & YFCCv   && -- & $76.4^\dagger$  && -- & -- \\
	  Wang and Gupta~\cite{wang2015unsupervised}      & YT8M        && -- & --        && -- & $60.2^\dagger$ \\
	  Wang~\etal~\cite{wang2017transitive}            & YT9M        && $59.4^{\phantom{\dagger}}$   & $79.6^{\phantom{\dagger}}$          && $40.9^{\phantom{\dagger}}$ & $63.2^\dagger$ \\
    \midrule
    \OURS                                           & YFCC         && $\mathbf{79.7}^{\phantom{\dagger}}$ & $\mathbf{84.3}^{\phantom{\dagger}}$ && $\mathbf{60.5}^{\phantom{\dagger}}$ & $\mathbf{67.8}^{\phantom{\dagger}}$ \\
    \bottomrule
  \end{tabular}
  \caption{
Comparison of \OURS to state-of-the-art unsupervised feature learning on classification and detection on \textsc{Pascal} VOC $2007$.
We disassociate methods using curated datasets and methods using non-curated datasets.
We selected hyper-parameters for each transfer task on the validation set, and then retrain on both training and validation sets.
We report results on the test set averaged over $5$ runs.
``YFFCv'' stands for the videos contained in YFFC100M dataset.
$^\dagger$ numbers from their original paper.
}
  \label{tab:voc}
\end{table}

\paragraph{Pascal VOC 2007~\cite{everingham2010pascal}.}
This dataset has small training and validation sets ($2.5$k images each), making it close to the setting of real applications where models trained using large computational resources are adapted to a new task with a small number of instances.
We report numbers on the classification and detection tasks with finetuning (``\textsc{all}'') or by only retraining the last three fully connected layers of the network (``\textsc{fc68}'').
The \textsc{fc68} setting gives a better measure of the quality of the evaluated features since fewer parameters are retrained.
For classification, we use the code of Caron~\etal~\cite{caron2018deep}\footnote{\scriptsize\url{github.com/facebookresearch/deepcluster}} and for detection, \texttt{fast-rcnn}~\cite{girshick2015fast}\footnote{\scriptsize\url{github.com/rbgirshick/py-faster-rcnn}}.
For classification, we train the models for $150k$ iterations, starting with a learning rate of $0.002$ decayed by a factor $10$ every $20k$ iterations, and we report results averaged over $10$ random crops.
For object detection, we train our network for $150k$ iterations, dividing the step-size by $10$ after the first $50k$ steps with an initial learning rate of $0.01$ (\textsc{fc68}) or $0.002$ (\textsc{all}) and a weight decay of $0.0001$.
Following Doersch~\etal~\cite{doersch2015unsupervised}, we use the multiscale configuration, with scales $[400,500,600,700]$ for training and $[400,500,600]$ for testing.
In Table~\ref{tab:voc}, we compare \OURS with two sets of unsupervised methods that use a VGG-16 network: those trained on curated datasets and those trained on non-curated datasets.
Previous unsupervised methods that worked on unucurated datasets with a VGG-16 use videos: Youtube8M (``YT8M''), Youtube9M (``YT9M'') or the videos from YFCC100M (``YFFCv'').
Our approach achieves state-of-the-art performance among all the unsupervised method that uses a VGG-16 architecture, even those that use ImageNet as a training set.
The gap with a supervised network is still important when we freeze the convolutions ($6\%$ for detection and $10\%$ for classification) but drops to less than $5\%$ for both tasks with finetuning.

\begin{figure}[t]
	\centering
  \includegraphics[width=0.48\linewidth]{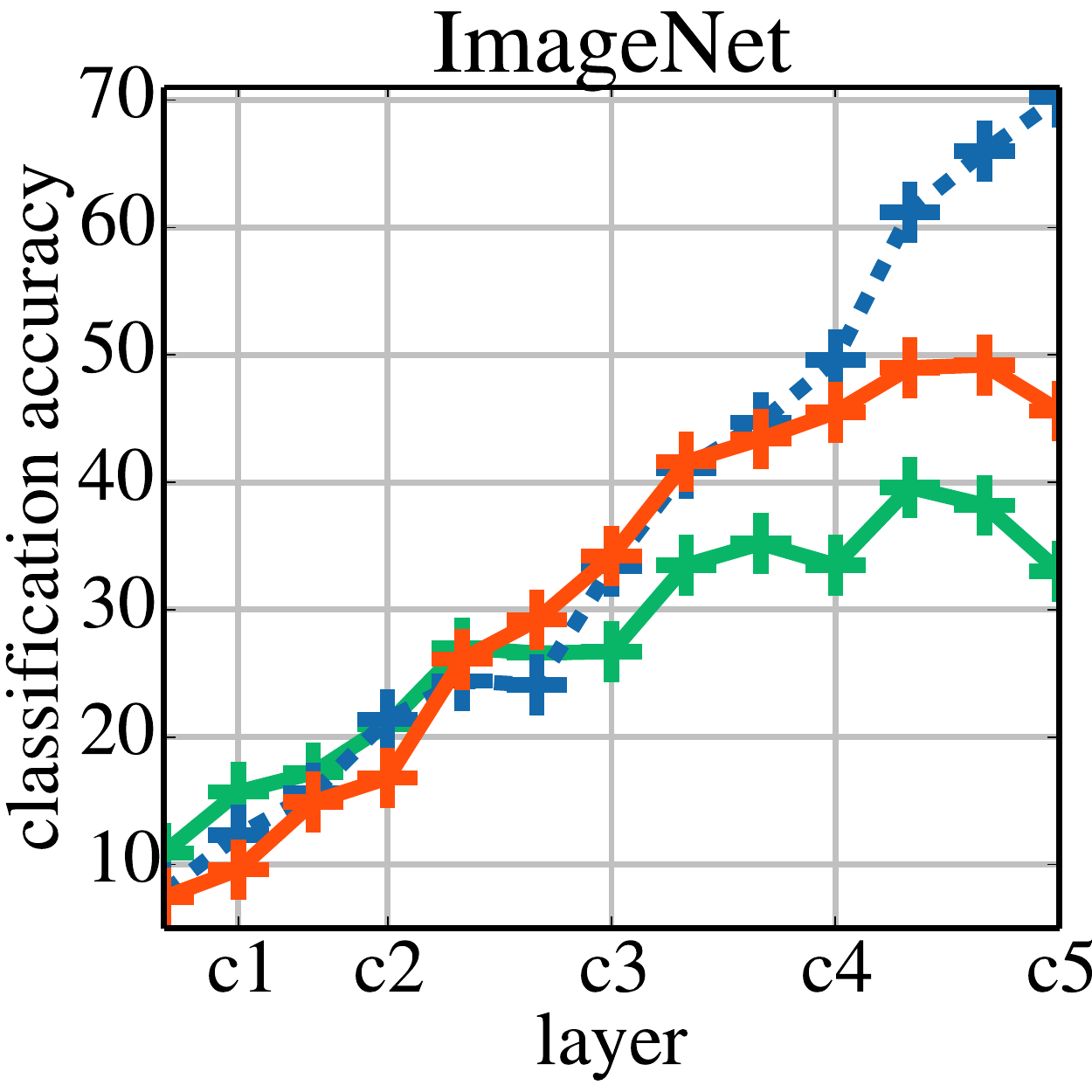}
  \includegraphics[width=0.48\linewidth]{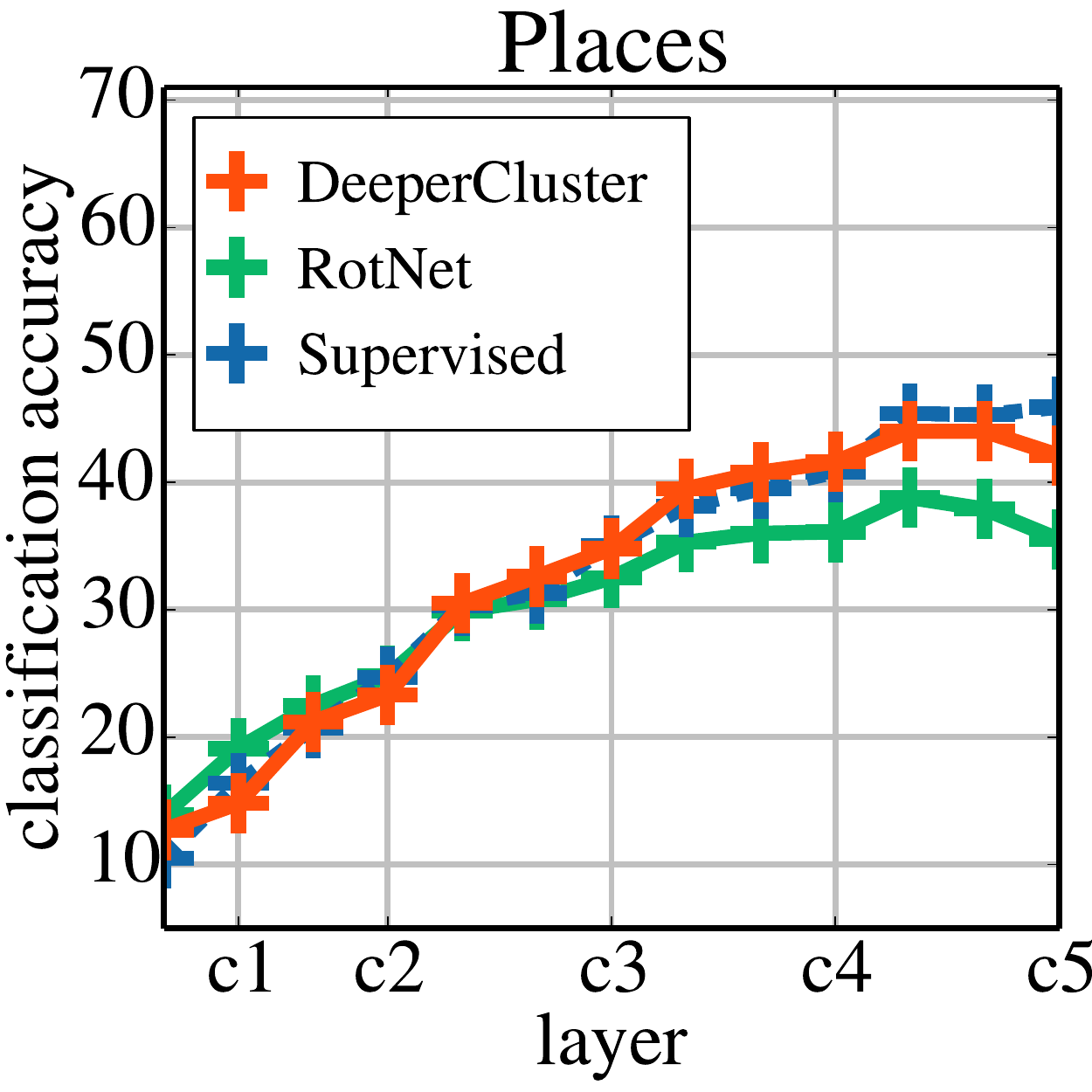}
	\caption{
Accuracy of linear classifiers on ImageNet and Places205 using the activations from different layers as features.
We compare a VGG-16 trained with supervision on ImageNet to VGG-16 trained with either RotNet or \OURS on YFCC100M.
Exact numbers are in Appendix.
	}
	\label{fig:layers}
\end{figure}

\paragraph{Linear classifiers on ImageNet~\cite{deng2009imagenet} and Places205~\cite{zhou2014learning}.}
ImageNet (``INet'') and Places205 (``Pl.'') are two large scale image classification datasets: ImageNet's domain covers objects and animals ($1.3$M images) and Places205's domain covers indoor and outdoor scenes ($2.5$M images).
We train linear classifiers with a logistic loss on top of frozen convolutional layers at different depths.
To reduce influence of feature dimension in the comparison, we average-pool the features until their dimension is below $10k$~\cite{zhang2016colorful}.
This experiment probes the quality of the features extracted at each convolutional layer.
In Figure~\ref{fig:layers}, we observe that \OURS matches the performance of a supervised network for all layers on Places205.
On ImageNet, it also matches supervised features up to the $4$th convolutional block; then the gap suddenly increases to around $20\%$.
It is not surprising since the supervised features are trained on ImageNet itself, while ours are trained on YFCC100M.

\subsection{Pre-training for ImageNet}
In the previous section, we can observe that a VGG-16 trained on YFCC100M has similar or better low level features than the same network trained on ImageNet with supervision.
In this experiment, we want to check whether these low-level features pre-trained on YFCC100M without supervision can serve as a good initialization for fully-supervised ImageNet classification.
To this end, we pre-train a VGG-16 on YFCC100M using either \OURS or RotNet.
The resulting weights are then used as initialization for the training of a network on ImageNet with supervision.
We merge the Sobel weights of the network pre-trained with \OURS with the first convolutional layer during the initialization.
We then train the networks on ImageNet with mini-batch SGD for $100$ epochs, a learning rate of $0.1$, a weight decay of $0.0001$, a batch size of $256$ and dropout of $0.5$.
We reduce the learning rate by a factor of $0.2$ every $20$ epochs.
Note that this learning rate decay schedule slightly differs from the ImageNet classification PyTorch default implementation\footnote{\scriptsize\url{github.com/pytorch/examples/blob/master/imagenet/}} where they train for $90$ epochs and decay the learning rate by $0.1$ at epochs $30$ and $60$.
We give in Appendix the results with this default schedule (with unchanged conclusions).
In Table~\ref{tab:pretrain}, we compare the performance of a network trained with a standard intialization (``Supervised'') to one initialized with a pre-training obtained from either \OURS (``Supervised + \OURS pre-training'') or RotNet (``Supervised + RotNet pre-training'') on YFCC100M.
We see that our pre-training improves the performance of a supervised network by $+0.8\%$, leading to $74.9\%$ top-1 accuracy.
This means that our pre-training captures important statistics from YFCC100M that transfers well to ImageNet.

\begin{table}[t!]
\centering
  \setlength{\tabcolsep}{3.5pt}
    \begin{tabular}{@{}l c cc @{}}
      \toprule
      ImageNet &~~~&top-$1$ & top-$5$ \\
      \midrule
      Supervised (PyTorch documentation\footnote{\scriptsize\url{pytorch.org/docs/stable/torchvision/models}}) && $73.4$ & $91.5$ \\
      Supervised (our code)  && $74.1$ & $91.8$ \\
      Supervised + RotNet pre-training && $74.5$ & $92.0$ \\
      Supervised + \OURS pre-training && $\mathbf{74.9}$ & $\mathbf{92.3}$ \\
      \bottomrule
    \end{tabular}
    \caption{
      Accuracy on the validation set of ImageNet classification for a supervised VGG-16 trained with different initializations:
      we compare a network trained from a standard initialization to networks trained from pre-trained weights using either \OURS or RotNet on YFCC100M.
    }
    \label{tab:pretrain}
\end{table}

\subsection{Model analysis}
In this final set of experiments, we analyze some components of our model.
Since \OURS derives from RotNet and DeepCluster,
we first look at the difference between these methods and ours, when trained on curated and non-curated datasets.
We then report quantitative and qualitative evaluations of the clusters obtained with \OURS.

\paragraph{Comparison with RotNet and DeepCluster.}
In Table~\ref{tab:linear}, we compare \OURS with DeepCluster and RotNet when a linear classifier is trained on top of the last convolutional layer of a VGG-16 on several datasets.
For reference, we also report previously published numbers~\cite{wu2018unsupervised} with a VGG-$16$ architecture.
We average-pool the features of the last layer resulting in representations of $8192$ dimensions.
Our approach outperforms both RotNet and DeepCluster, even when they are trained on curated datasets (except for ImageNet classification task where DeepCluster trained on ImageNet yields the best performance).
More interestingly, we see that the quality of the dataset or its scale has little impact on RotNet while it has on DeepCluster.
This is confirming that self-supervised methods are more robust than clustering to a change of dataset distribution.

\begin{table}[t!]
\centering
  \setlength{\tabcolsep}{3.5pt}
    \begin{tabular}{@{}l c c c c@{}}
      \toprule
      Method & Data & ImageNet & Places & VOC2007 \\
      \midrule
      Supervised  & ImageNet  & $70.2$ & $45.9$  & $84.8$ \\
      \midrule
      Wu~\etal~\cite{wu2018unsupervised} & ImageNet & $39.2$ & $36.3$ & - \\
      \midrule
      RotNet      & ImageNet & $32.7$ & $32.6$ &  $60.9$ \\
      DeepCluster & ImageNet & $\mathbf{48.4}$ & $37.9$ &  $71.9$ \\
      \midrule
      RotNet      & YFCC100M & $33.0$ & $35.5$ &  $62.2$ \\
      DeepCluster & YFCC100M & $34.1$ & $35.4$ &  $63.9$ \\
      \midrule
      \OURS & YFCC100M & $45.6$ & $\mathbf{42.1}$  &  $\mathbf{73.0}$ \\
      \bottomrule
    \end{tabular}
    \caption{
      Comparaison between \OURS, RotNet and DeepCluster when pre-trained on curated and non-curated dataset.
      We report the accuracy on several datasets of a linear classifier trained on top of features of the last convolutional layer.
      All the methods use the same architecture.
      DeepCluster does not scale to the full YFCC100M dataset, we thus train it on a random subset of $1.3$M images.
    }
    \label{tab:linear}
\end{table}

\begin{figure*}[t]
	\centering
    \includegraphics[height=0.19\linewidth]{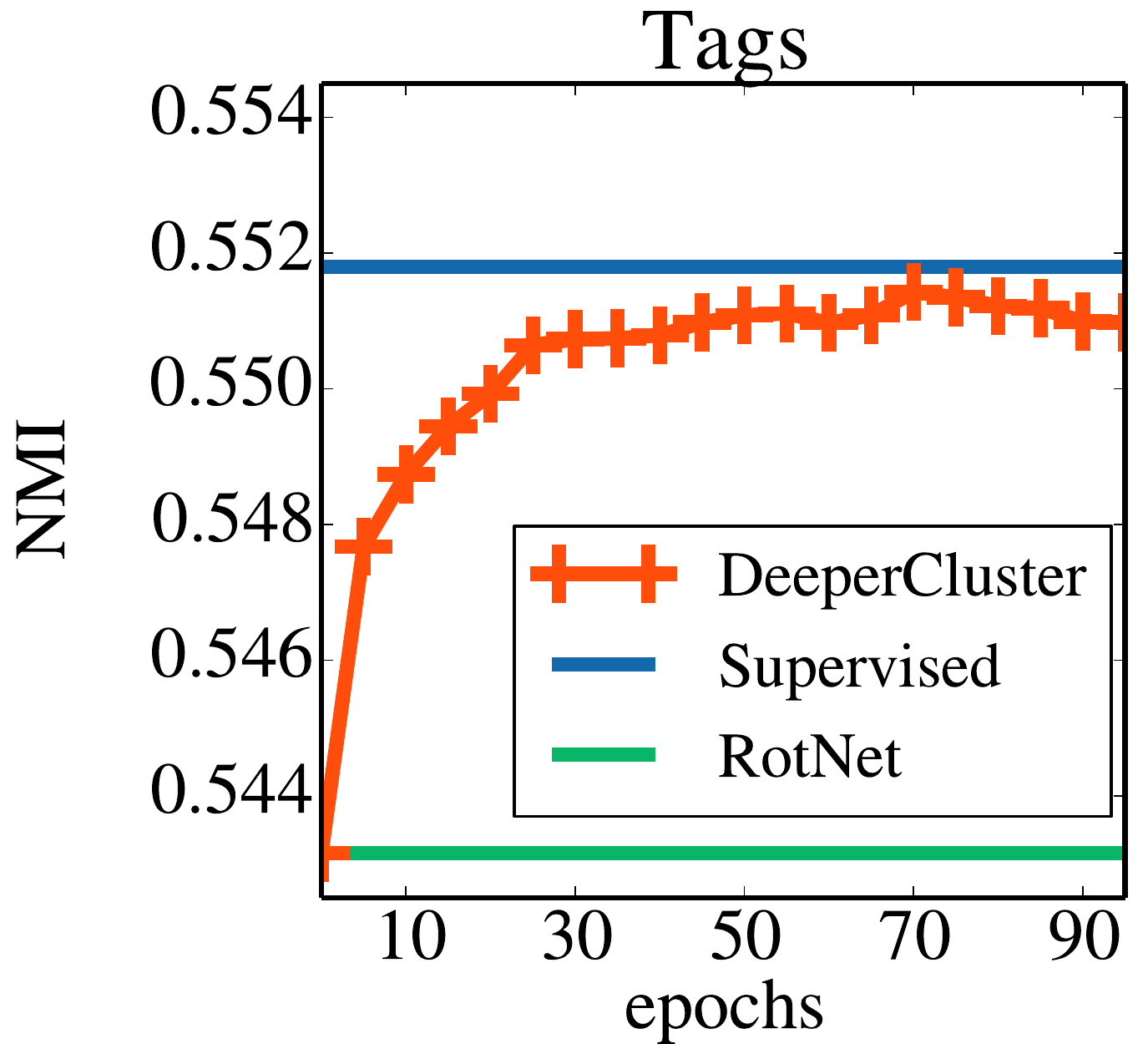}
    \includegraphics[height=0.19\linewidth]{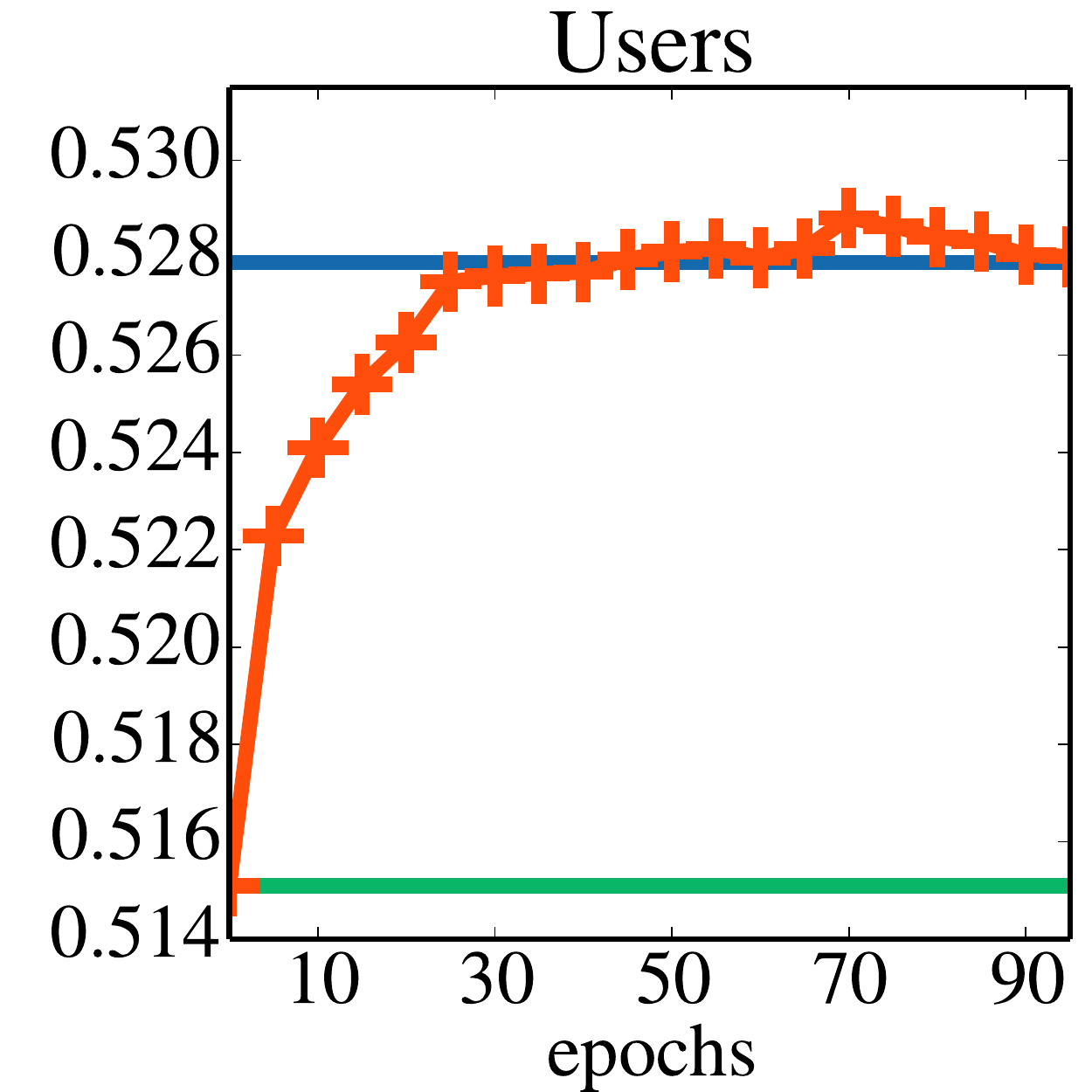}
    \includegraphics[height=0.19\linewidth]{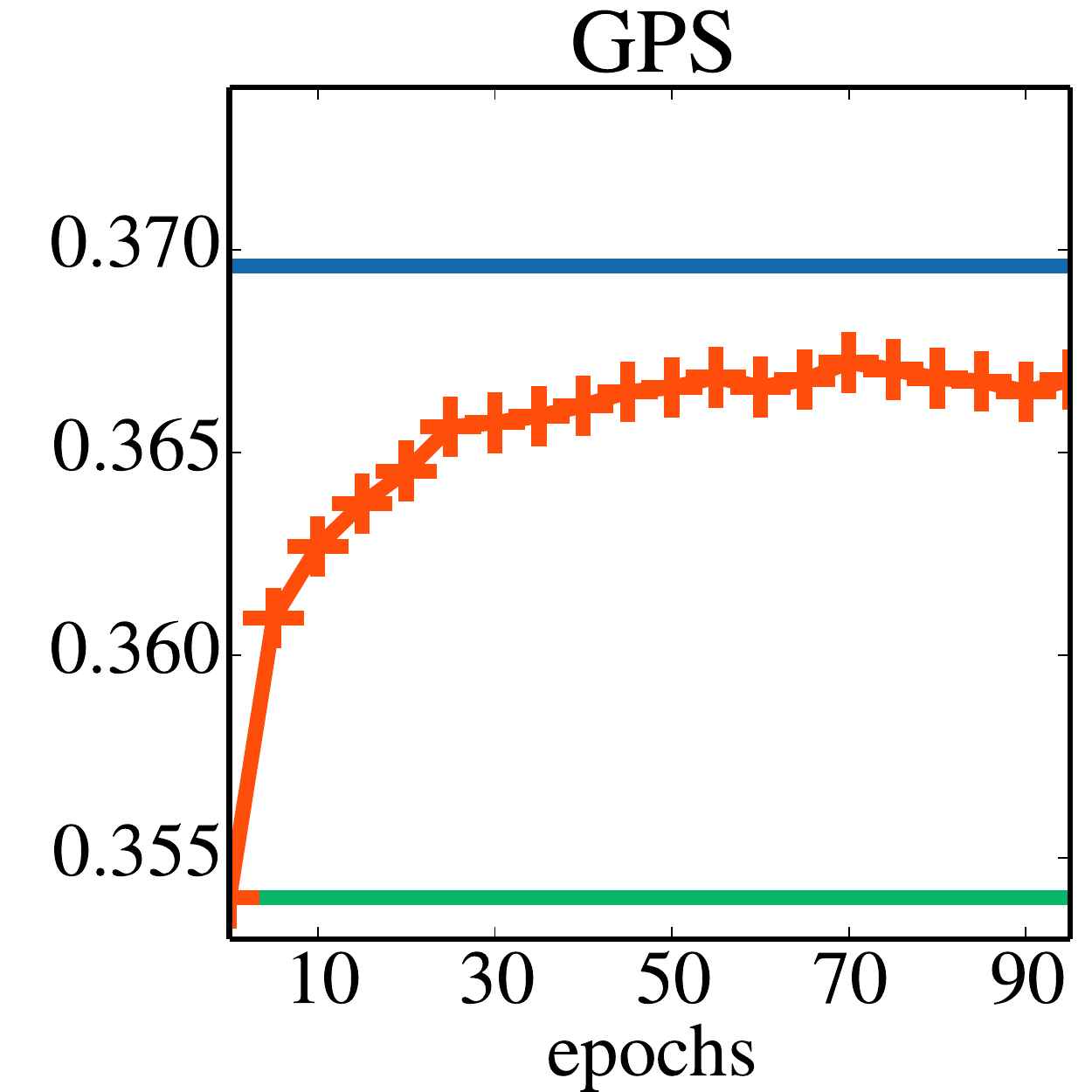}
    \includegraphics[height=0.19\linewidth]{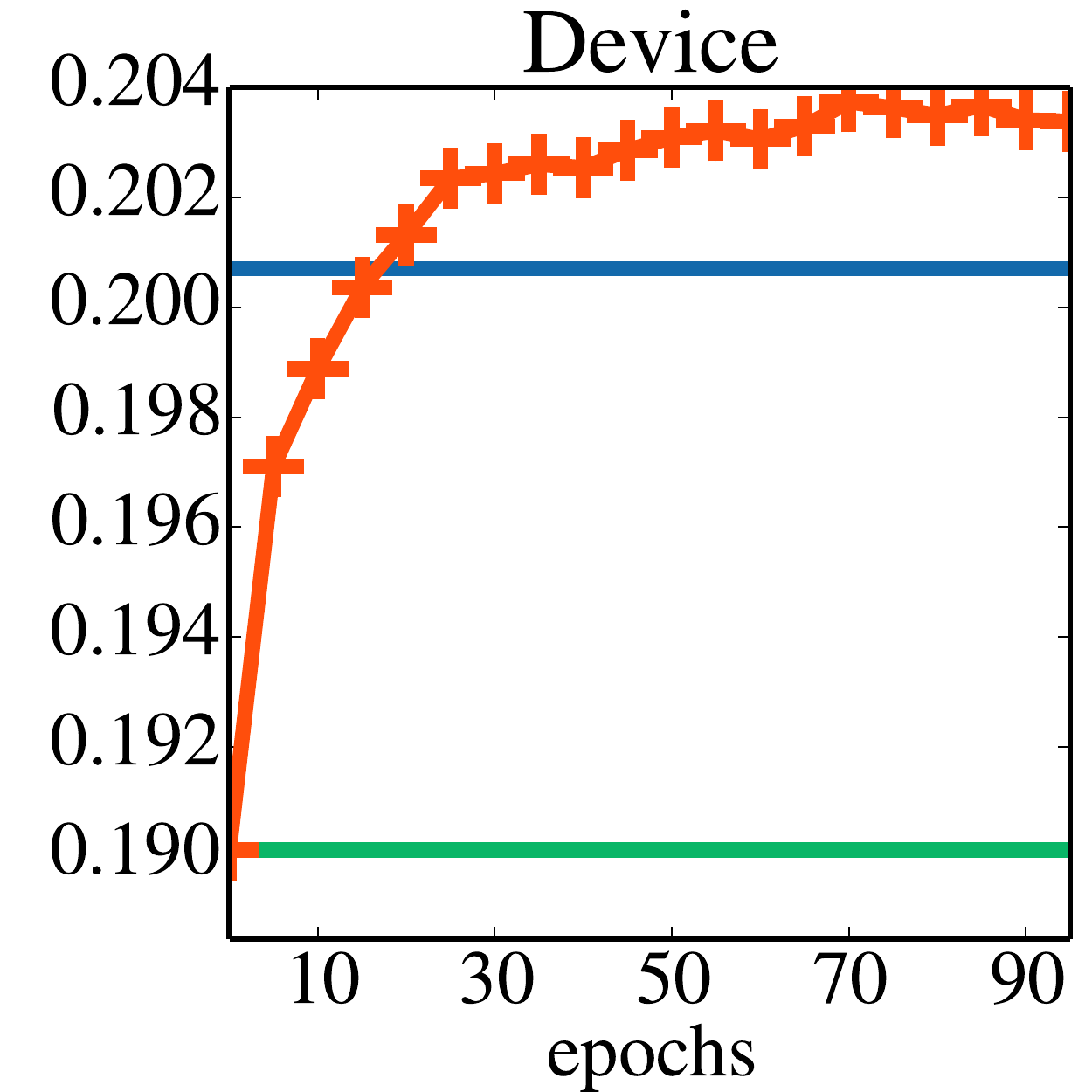}
    \includegraphics[height=0.19\linewidth]{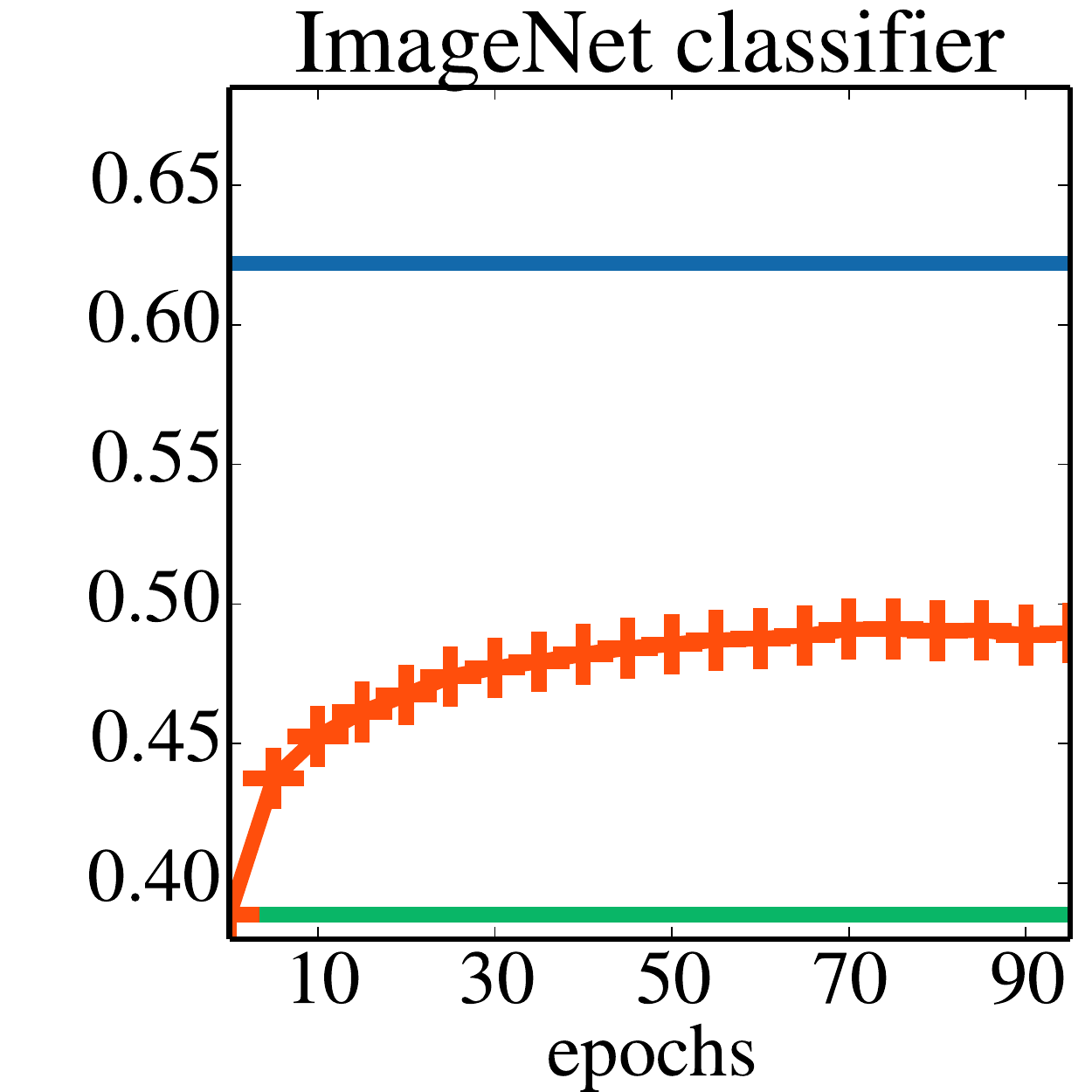}
  \caption{Normalized mutual information between our clustering and different sorts of metadata: hashtags, user IDs, geographic coordinates, and device types.
   We also plot the NMI with an ImageNet classifier labeling.}
  \label{fig:nmi}
\end{figure*}

\begin{figure*}[t!]
  \centering
  \begin{tabular}{ccccccc}
	  {tag: \itshape cat} & {tag: \itshape elephantparadelondon} & {tag: \itshape always} & {device: \itshape CanoScan}
  \\
  \includegraphics[width=0.225\linewidth]{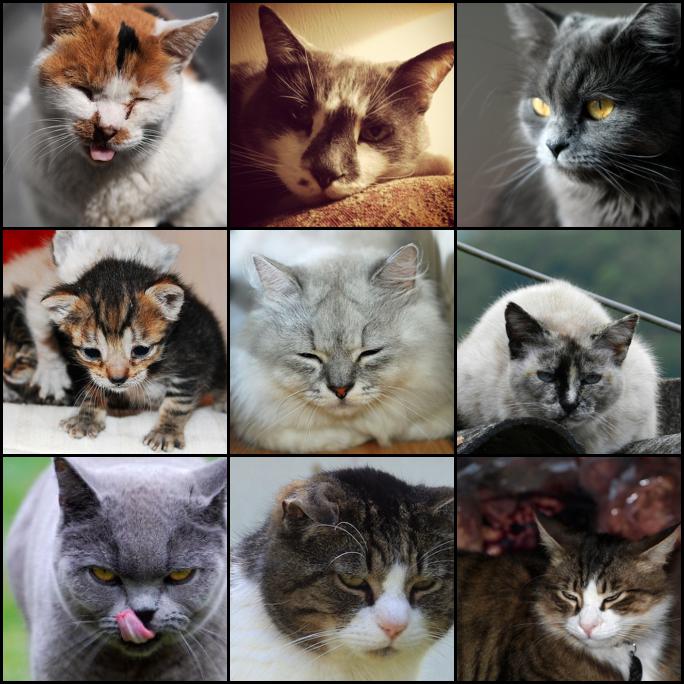}&
  \includegraphics[width=0.225\linewidth]{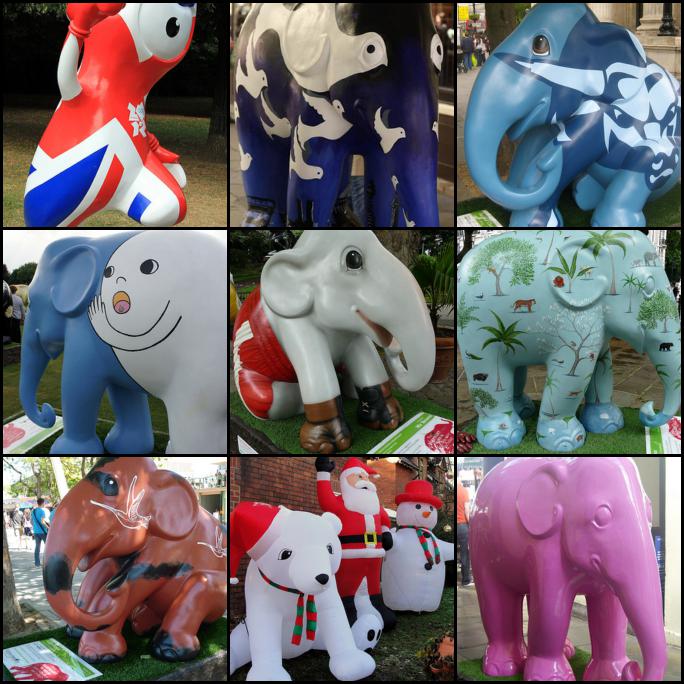}&
  \includegraphics[width=0.225\linewidth]{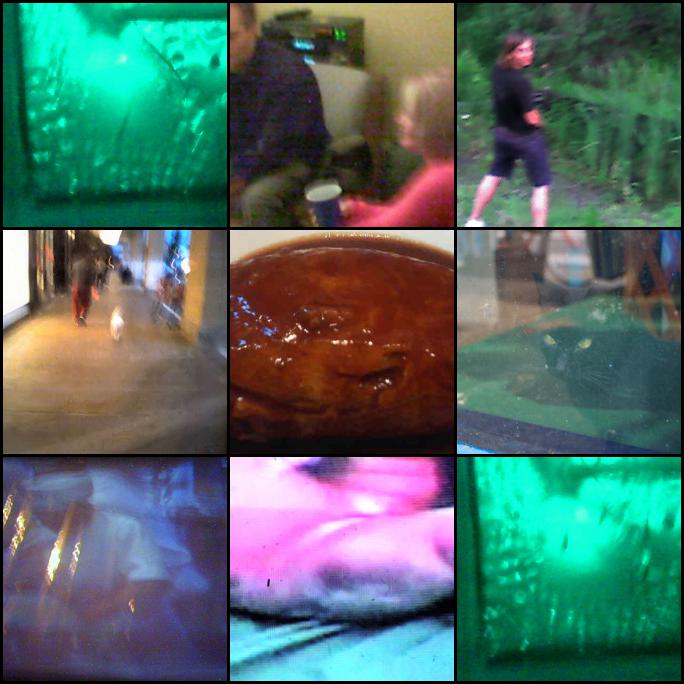}&
  \includegraphics[width=0.225\linewidth]{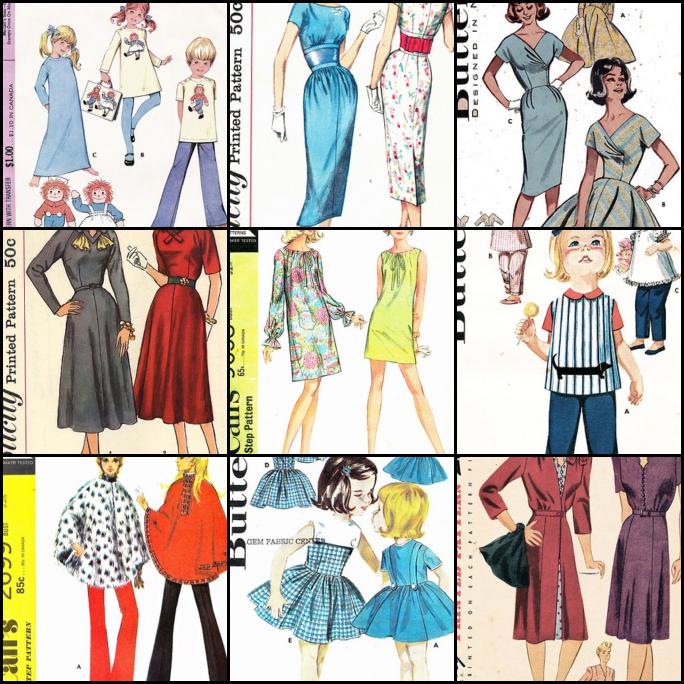}
  \\
  {GPS: ($43$, $10$)} & {GPS: ($-34$, $-151$)} & {GPS: ($64$, $-20$)} & {GPS: ($43$, $-104$)}
  \\
  \includegraphics[width=0.225\linewidth]{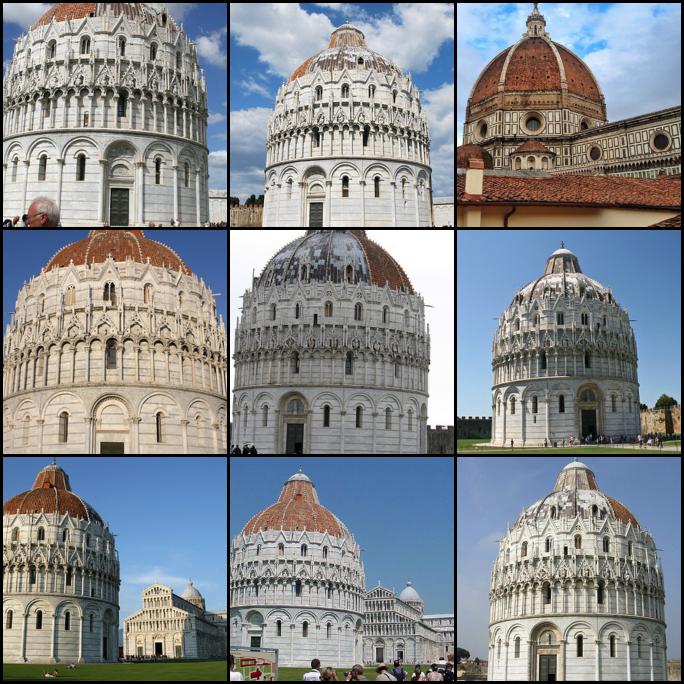}&
  \includegraphics[width=0.225\linewidth]{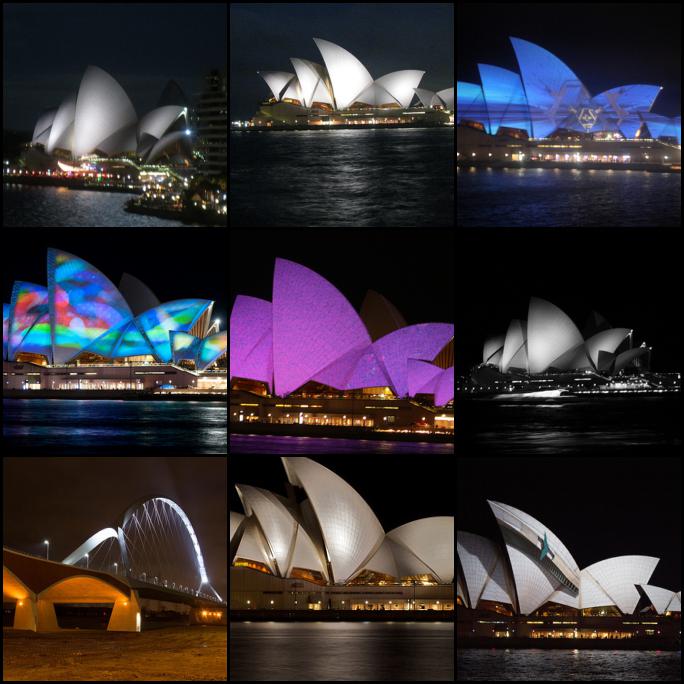}&
  \includegraphics[width=0.225\linewidth]{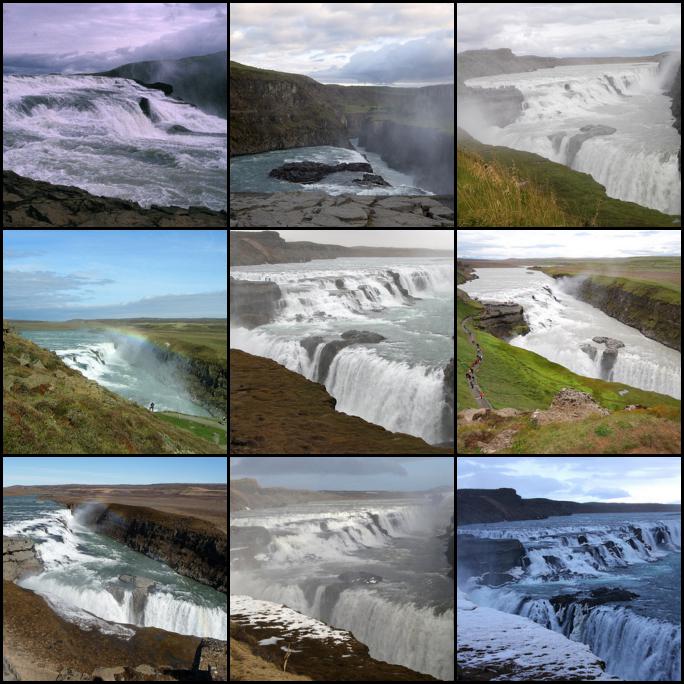}&
  \includegraphics[width=0.225\linewidth]{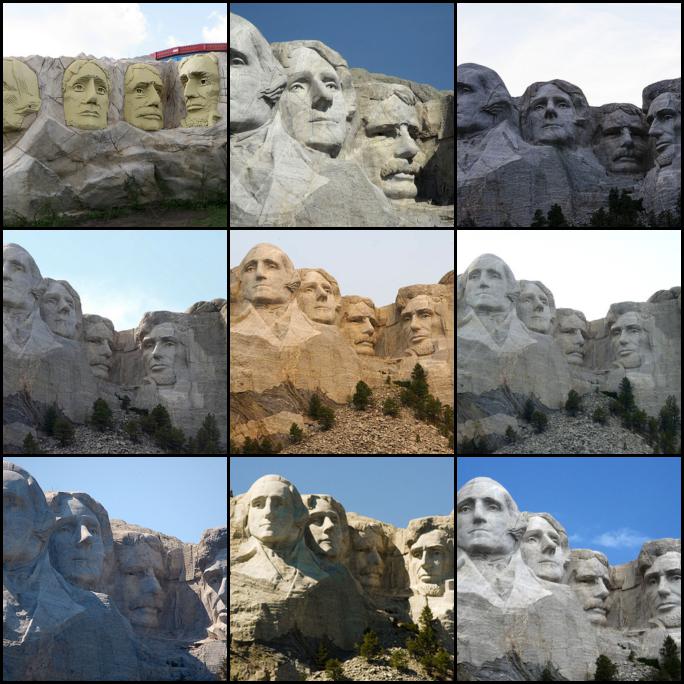}
  \end{tabular}
  \caption{We randomly select $9$ images per cluster and indicate the dominant cluster metadata.
The bottom row depicts clusters pure for GPS coordinates but unpure for user IDs.
As expected, they turn out to correlate with tourist landmarks.
No metadata is used during training.
For copyright reasons, we provide in Appendix the photographer username for each image.
  }
  \label{fig:cluster}
\end{figure*}

\paragraph{Influence of dataset size and number of clusters.}
\label{sec:exp-k}
To measure the influence of the number of images on features, we train models with $1$M, $4$M, $20$M, and $96$M images and report their accuracy on the validation set of the Pascal VOC 2007 classification task (\textsc{fc68} setting).
We also train models on $20$M images with a number of clusters that varies from $10$k to $160$k.
For the experiment with a total of $160$k clusters, we choose $m=2$ which results in $8$ super-classes.
In Figure~\ref{fig:k}, we observe that the quality of our features improves when scaling both in terms of images and clusters.
Interestingly, between $4$M and $20$M of YFCC100M images are needed to meet the performance of our method on ImageNet.
Augmenting the number of images has a bigger impact than the number of clusters.
Yet, this improvement is significant since it corresponds to a reduction of more than $10\%$ of the relative error w.r.t. the supervised model.

\paragraph{Quality of the clusters.}
In addition to features, our method provides a clustering of the input images.
We evaluate the quality of these clusters by measuring their correlation with existing partitions of the data.
In particular, YFCC100M comes with many different metadata.
We consider hashtags, users, camera and GPS coordinates.
If an image has several hashtags, we pick as label the least frequent one in the total hashtag distribution.
We also measure the correlation of ours clusters with labels predicted by a classifier trained on ImageNet categories.
We use a ResNet-$50$ network~\cite{he2016deep}, pre-trained on ImageNet, to classify the YFCC100M images and we select those for which the confidence in prediction is higher than $75\%$.
This evaluation omits a large amount of the data but gives some insight about the quality of our clustering in object classification.

In Figure~\ref{fig:nmi}, we show the evolution during training of the normalized mutual information (NMI) between our clustering and different metadata, and the predicted labels from ImageNet.
The higher the NMI, the more correlated our clusters are to the considered partition.
For reference, we compute the NMI for a clustering of RotNet features (as it corresponds to weights at initialization) and of a supervised model.
First, it is interesting to observe that our clustering is improving over time for every type of metadata.
One important factor is that most of these commodities are correlated since a given user takes pictures in specific places with probably a single camera and use a preferred fixed set of hashtags.
Yet, these plots show that our model captures in the input signal enough information to predict these metadata at least as well as the features trained with supervision.

We visually assess the consistency of our clusters in Figure~\ref{fig:cluster}.
We display $9$ random images from $8$ manually picked clusters.
The first two clusters contain a majority of images associated with tag from the head (first cluster) and from the tail (second cluster) in the YFC100M dataset.
Indeed, $418.538$ YFC100M images are associated with the tag \textit{cat} whereas only $384$ images contain the tag \textit{elephantparadelondon} ($0.0004\%$ of the dataset).
We also show a cluster for which the dominant hashtag does not corrolate visually with the content of the cluster.
As already mentioned, this database is non-curated and contains images that basically do not depict anything semantic.
The dominant metadata of the last cluster in the top row is the device ID \textit{CanoScan}.
As this cluster is about drawings, its images have been mainly taken with a scanner.
Finally, the bottom row depict clusters that are pure for GPS coordinates but unpure for user IDs.
It results in clusters of images taken by many different users in the same place: tourist landmarks.

\section{Conclusion}
In this paper, we present an unsupervised approach specifically designed to deal with large amount of non-curated data.
Our method is well-suited for distributed training, which allows training on large datasets with $96M$ of images.
With such amount of data, our approach surpasses unsupervised methods trained on curated datasets, which validates the potential of unsupervised learning in applications where annotations are scarce or curation is not trivial.
Finally, we show that unsupervised pre-training improves the performance of a network trained on ImageNet.

\paragraph{Acknowledgement.}
We thank Thomas Lucas, Matthijs Douze, Francisco Massa and the rest of Thoth and FAIR teams for their help and fruitful discussions.
We also thank the anonymous reviewers for their thoughtful feedback.
Julien Mairal was funded by the ERC grant number 714381 (SOLARIS project).

{\small
\bibliographystyle{ieee_fullname}
\bibliography{egbib}
}

\clearpage
\section*{\textbf{\LARGE{Appendix}}}

\begin{table*}[h]
\centering
  \setlength{\tabcolsep}{3.5pt}
    \begin{tabular}{@{}l ccccccccccccc@{}}
      \toprule
      Method & \footnotesize \texttt{conv1} & \footnotesize \texttt{conv2} & \footnotesize \texttt{conv3} & \footnotesize \texttt{conv4} & \footnotesize \texttt{conv5} & \footnotesize \texttt{conv6} & \footnotesize \texttt{conv7} & \footnotesize \texttt{conv8} & \footnotesize \texttt{conv9} & \footnotesize \texttt{conv10} & \footnotesize \texttt{conv11} & \footnotesize \texttt{conv12} & \footnotesize \texttt{conv13} \\
      \midrule
      \textit{ImageNet} \\
      \midrule
      Supervised & $7.8 $ & $12.3$ & $15.6$ & $21.4$ & $24.4$ & $24.1$ & $33.4$ & $41.1$ & $44.7$ & $49.6$ & $61.2$ & $66.0$ & $70.2$ \\
      RotNet & $10.9$ & $15.7$ & $17.2$ & $21.0$ & $27.0$ & $26.6$ & $26.7$ & $33.5$ & $35.2$ & $33.5$ & $39.6$ & $38.2$ & $33.0$ \\
      \OURS & $7.4$ & $9.6$ & $14.9$ & $16.8$ & $26.1$ & $29.2$ & $34.2$ & $41.6$ & $43.4$ & $45.5$ & $49.0$ & $49.2$ & $45.6$ \\
      \midrule
      \textit{Places205} \\
      \midrule
      Supervised & $10.5$ & $16.4$ & $20.7$ & $24.7$ & $30.3$ & $31.3$ & $35.0$ & $38.1$ & $39.5$ & $40.8$ & $45.4$ & $45.3$ & $45.9$ \\
      RotNet & $13.9$ & $19.1$ & $22.5$ & $24.8$ & $29.9$ & $30.8$ & $32.5$ & $35.3$ & $36.0$ & $36.1$ & $38.8$ & $37.9$ & $35.5$ \\
      \OURS & $12.7$ & $14.8$ & $21.2$ & $23.3$ & $30.5$ & $32.6$ & $34.8$ & $39.5$ & $40.8$ & $41.6$ & $44.0$ & $44.0$ & $42.1$ \\
      \bottomrule
    \end{tabular}
    \caption{
    Accuracy of linear classifiers on ImageNet and Places205 using the activations from different layers as features.
    We train a linear classifier on top of frozen convolutional  layers at different depths.
    We compare a VGG-16 trained with supervision on ImageNet to VGG-16s trained with either RotNet or our approach on YFCC100M.
    }
    \label{tab:layers}
\end{table*}

\setcounter{section}{0}

\section{Evaluating unsupervised features}
Here we provide numbers from Figure~$2$ in Table~\ref{tab:layers}.

\section{YFCC100M and Imagenet label distribution}

\begin{figure}[t]
  \centering
  \includegraphics[scale=0.3]{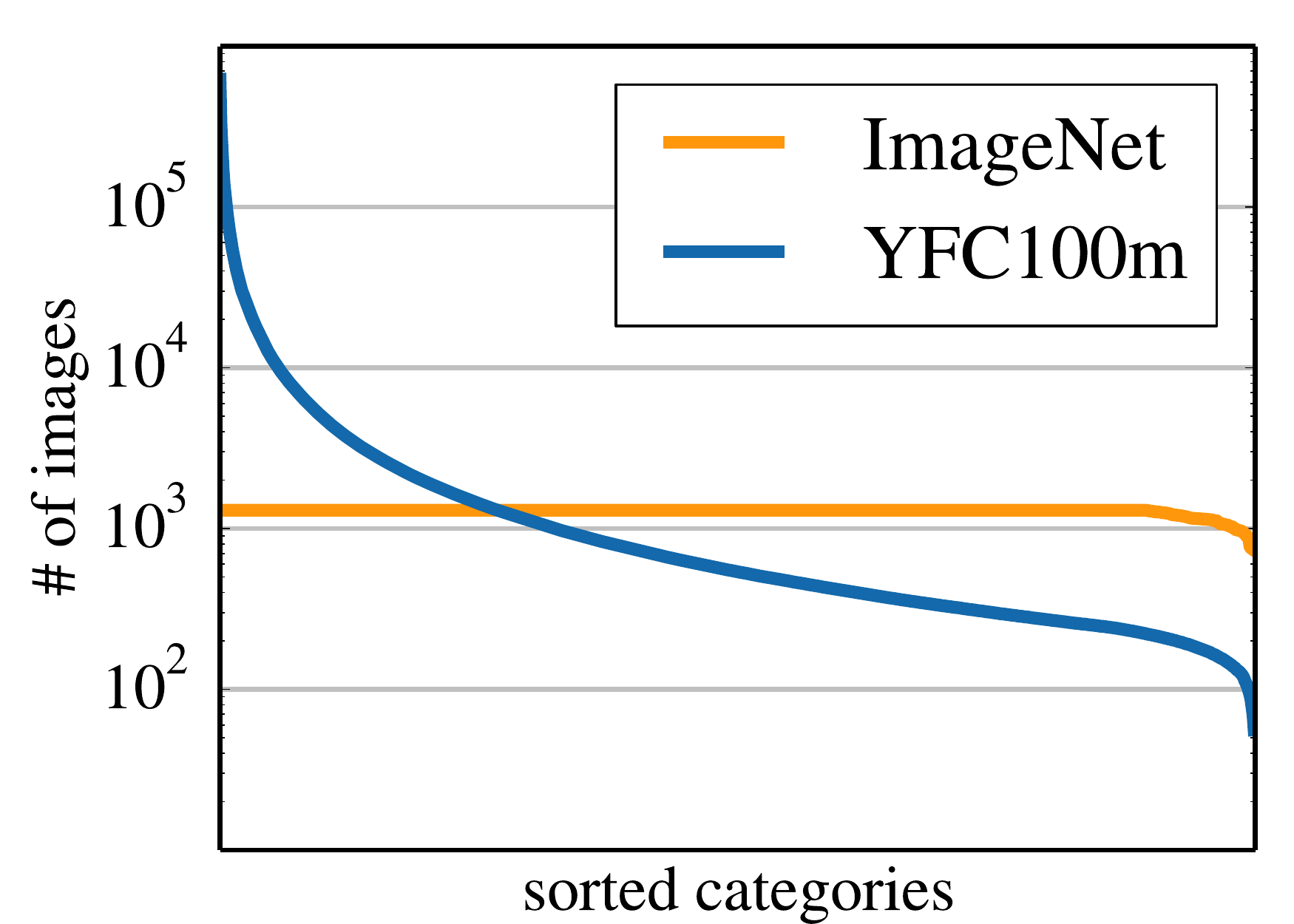}
  \caption{
    Comparison of the hashtag distribution in YFCC100M with the label distribution in ImageNet.
  }
  \label{fig:tag_dist}
\end{figure}

YFCC100M dataset contains social media from the Flickr website.
The content of this dataset is very unbalanced, with a ``long-tail'' distribution of hashtags contrasting with the well-behaved label distribution of ImageNet as can be seen in Figure~\ref{fig:tag_dist}.
For example, \textit{guenon} and \textit{baseball} correspond to labels with $1300$ associated images in ImageNet, while there are respectively $226$ and $256,758$ images associated with these hashtags in YFCC100M.

\section{Pre-training for ImageNet}

\begin{table}[h]
\centering
  \setlength{\tabcolsep}{3.5pt}
    \begin{tabular}{@{}l cc @{}}
      \toprule
	    & \footnotesize PyTorch doc & \footnotesize Our \\
	    & \footnotesize hyperparam & \footnotesize hyperparam \\
      \midrule
      	    \small Supervised (PyTorch documentation\footnote{\url{pytorch.org/docs/stable/torchvision/models}}) & $73.4$ & - \\
	    \small Supervised (our code)  & $73.3$ & $74.1$ \\
	    \small Supervised + RotNet pre-training & $73.7$ & $74.5$ \\
	    \small Supervised + \OURS pre-training & $74.3$ & $74.9$ \\
      \bottomrule
    \end{tabular}
    \caption{
      Top-$1$ accuracy on validation set of a VGG-16 trained on ImageNet with supervision with different initializations.
      We compare a network initialized randomly to networks pre-trained with our unsupervised method or with RotNet on YFCC100M.
    }
    \label{tab:pretrain90}
\end{table}
In Table~\ref{tab:pretrain90}, we compare the performance of a network trained with supervision on ImageNet with a standard intialization (``Supervised'') to one pre-trained with \OURS (``Supervised + \OURS pre-training'') and to one pre-trained with RotNet (``Supervised + RotNet pre-training'').
The convnet is finetuned on ImageNet with supervision with mini-batch SGD following the hyperparameters of the ImageNet classification example implementation from PyTorch documentation\footnote{\url{github.com/pytorch/examples/blob/master/imagenet/main.py}}).
Indeed, we train for $90$ epochs (instead of $100$ epochs in Table~$3$ of the main paper).
We use a learning rate of $0.1$, a weight decay of $0.0001$, a batch size of $256$ and dropout of $0.5$.
We reduce the learning rate by a factor of $0.1$ at epochs $30$ and $60$ (instead of decaying the learning rate with a factor $0.2$ every $20$ epochs in Table~$3$ of the main paper).
This setting is unfair towards the supervised from scratch baseline since as we start the optimization with a good initialization we arrive at convergence earlier.
Indeed, we observe that the gap between our pretraining and the baseline shrinks from $1.0$ to $0.8$ when evaluating at convergence instead of evaluating before convergence.
As a matter of fact, the gap for the RotNet pretraining with the baseline remains the same: $0.4$.

\section{Model analysis}
\subsection{Instance retrieval}

\begin{table}[h]
\centering
    \begin{tabular}{@{}lc c c@{}}
      \toprule
	    Method & Pretraining & Oxford$5$K & Paris$6$K \\
      \midrule
      ImageNet labels & ImageNet  & $72.4$ & $81.5$  \\
      Random          & -         & $\phantom{0}6.9$  & $22.0$ \\
      \midrule
      Doersch~\etal~\cite{doersch2015unsupervised} & ImageNet & $35.4$ & $53.1$ \\
      Wang~\etal~\cite{wang2017transitive} & Youtube $9$M & $42.3$ & $58.0$ \\
      \midrule
      RotNet         & ImageNet & $48.2$ & $61.1$\\
      DeepCluster     & ImageNet & $\mathbf{61.1}$ & $\mathbf{74.9}$\\
      \midrule
      RotNet          & YFCC100M & $46.5$ & $59.2$ \\
      DeepCluster     & YFCC100M & $57.2$ & $74.6$ \\
      \midrule
      \OURS           & YFCC100M & $55.8$ & $73.4$ \\
      \bottomrule
    \end{tabular}
    \caption{
      mAP on instance-level image retrieval on Oxford and Paris dataset.
      We apply R-MAC with a resolution of $1024$ pixels and $3$ grid levels~\cite{tolias2015particular}.
      We disassociate the methods using unsupervised ImageNet and the methods using non-curated datasets.
      DeepCluster does not scale to the full YFCC100M dataset, we thus train it on a random subset of $1.3$M images.
    }
    \label{tab:retrieval}
\end{table}

Instance retrieval consists of retrieving from a corpus the most similar images to a given a query.
We follow the experimental setting of Tolias~\etal~\cite{tolias2015particular}:
we apply R-MAC with a resolution of $1024$ pixels and $3$ grid levels and we report mAP on instance-level image retrieval on Oxford Buildings~\cite{philbin2007object} and Paris~\cite{philbin2008lost} datasets.

As described by Dosovitskiy~\etal~\cite{dosovitskiy2016discriminative}, class-level supervision induces invariance to semantic categories.
This property may not be beneficial for other computer vision tasks such as instance-level recognition.
For that reason, descriptor matching and instance retrieval are tasks for which unsupervised feature learning might provide performance improvements.
Moreover, these tasks constitute evaluations that do not require any additionnal training step, allowing a straightforward comparison accross different methods.
We evaluate our method and compare it to previous work following the experimental setup proposed by Caron~\etal~\cite{caron2018deep}.
We report results for the instance retrieval task in Table~\ref{tab:retrieval}.

We observe that features trained with RotNet have significantly worse performance than DeepCluster both on Oxford$5$K and Paris$6$K.
This performance discrepancy means that properties acquired by classifying large rotations are not relevant to instance retrieval.
An explanation is that all images in Oxford$5$k and Paris$6$k have the same orientation as they picture buildings and landmarks.
As our method is a combination of the two paradigms, it suffers an important performance loss on Oxfork$5$K, but is not affected much on Paris$6$k.
These results emphasize the importance of having a diverse set of benchmarks to evaluate the quality of features produced by unsupervised learning methods.

\begin{figure}[ht]
\centering
\includegraphics[width=0.7\linewidth]{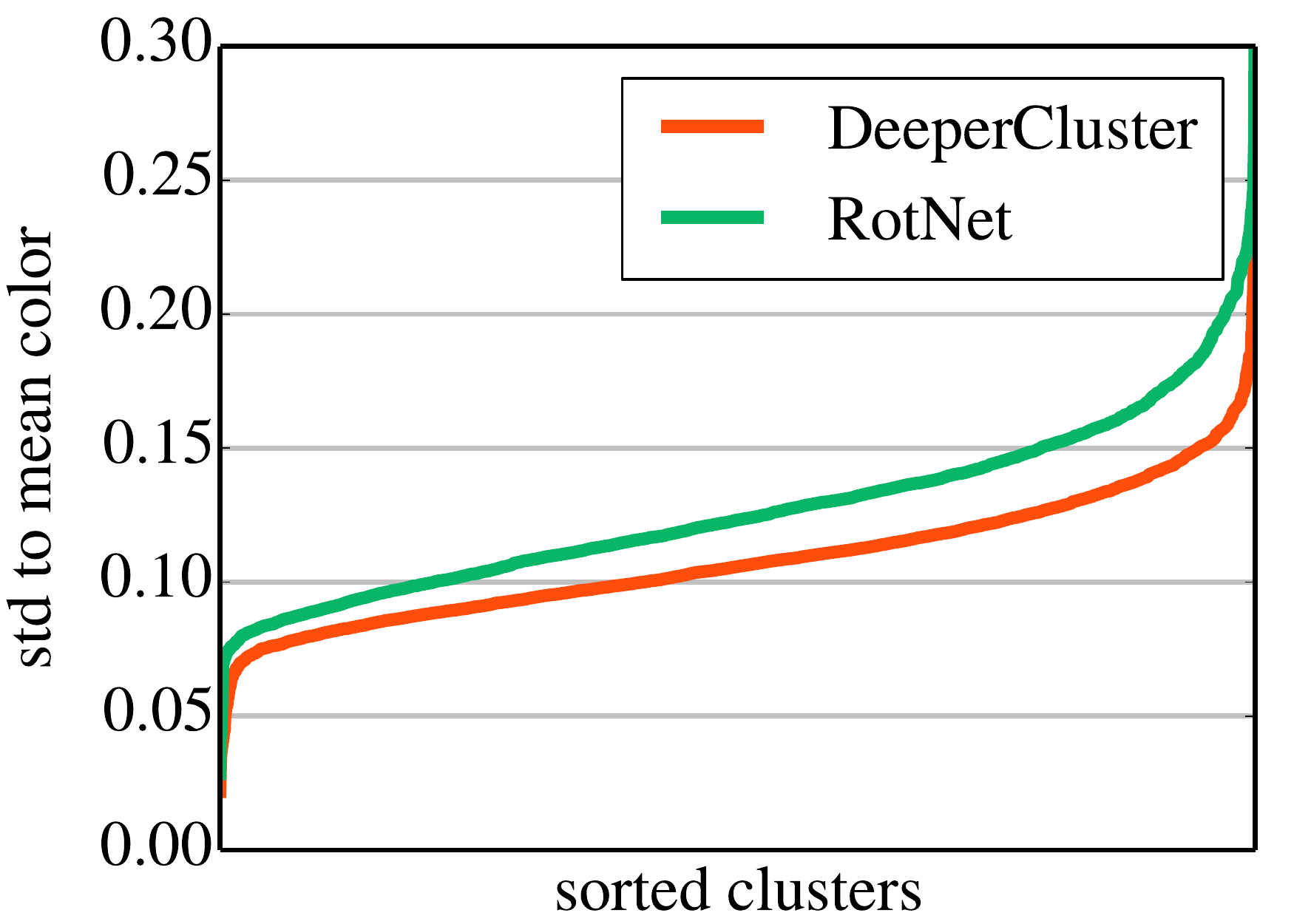}
\caption{Sorted standard deviations to clusters mean colors.
If the standard deviation of a cluster to its mean color is low, the images of this cluster have a similar colorization.}
\label{fig:color_std}
\end{figure}

\subsection{Influence of data pre-processing}
In this section we experiment with our method on raw RGB inputs.
We provide some insights into the reasons why sobel filtering is crucial to obtain good performance with our method.

\begin{figure}[ht]
\centering
  \begin{tabular}{cccc}
\includegraphics[width=0.2\linewidth]{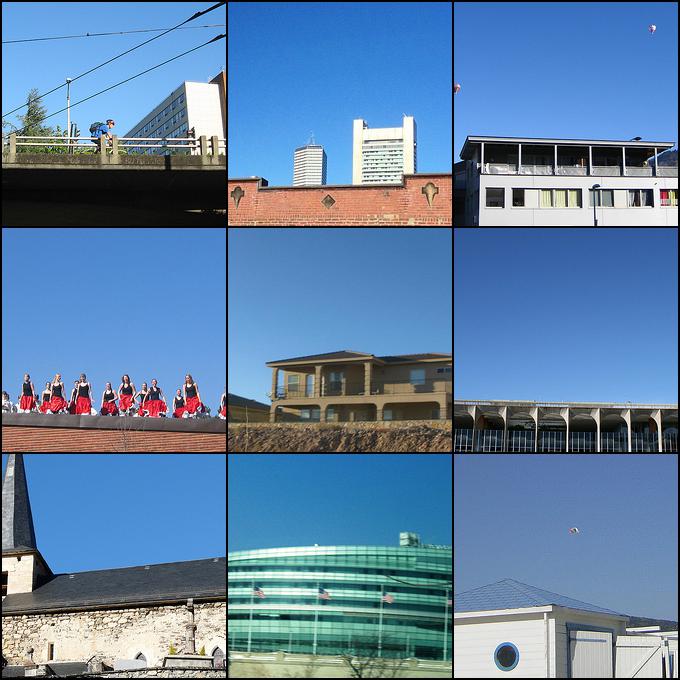}&
\includegraphics[width=0.2\linewidth]{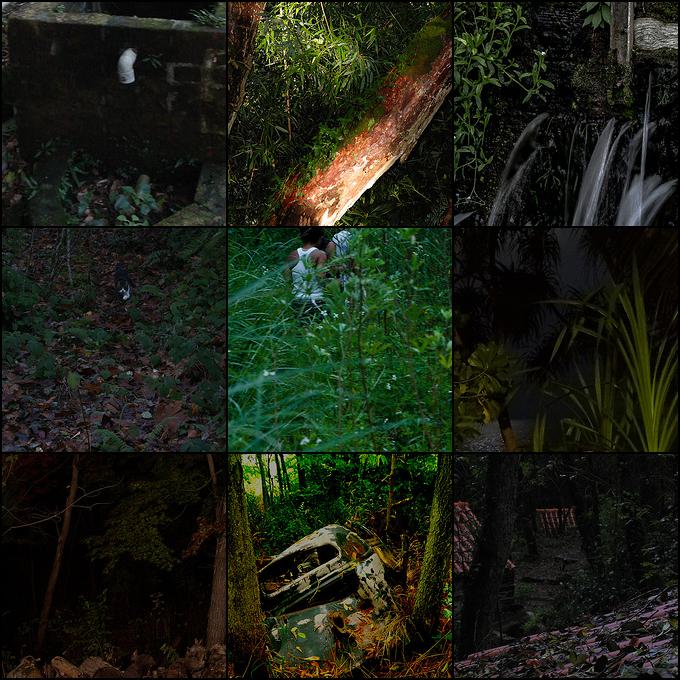}&
\includegraphics[width=0.2\linewidth]{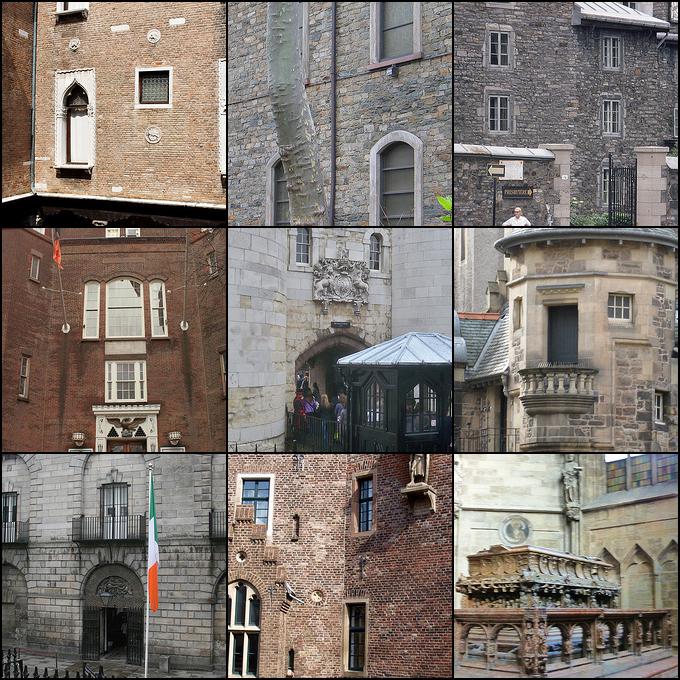}&
\includegraphics[width=0.2\linewidth]{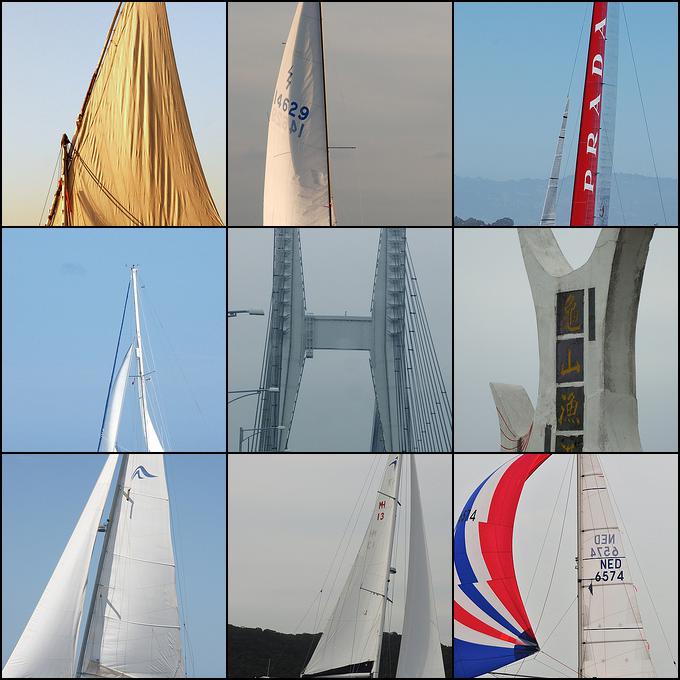}
\\
\\
\includegraphics[width=0.1\linewidth]{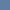}&
\includegraphics[width=0.1\linewidth]{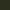}&
\includegraphics[width=0.1\linewidth]{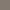}&
\includegraphics[width=0.1\linewidth]{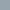}
\end{tabular}
\caption{We show clusters with an uniform colorization accross their images.
For each cluster, we show the mean color of the cluster.}
\label{fig:color_cluster}
\end{figure}

\begin{figure}[t]
\centering
  \begin{tabular}{c cc}
  RGB & \multicolumn{2}{c}{Sobel}\\
  \midrule
\includegraphics[width=0.27\linewidth]{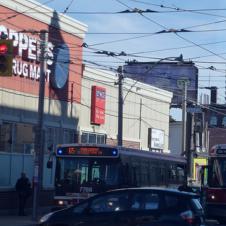}&
\includegraphics[width=0.27\linewidth]{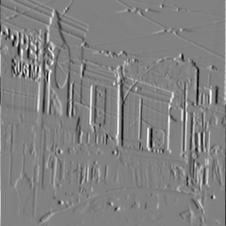}&
\includegraphics[width=0.27\linewidth]{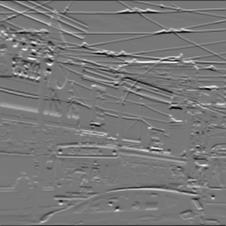}
\\
  \midrule
\includegraphics[width=0.27\linewidth]{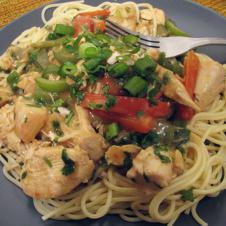}&
\includegraphics[width=0.27\linewidth]{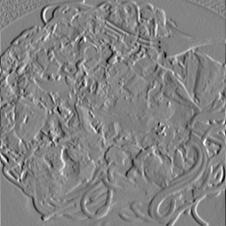}&
\includegraphics[width=0.27\linewidth]{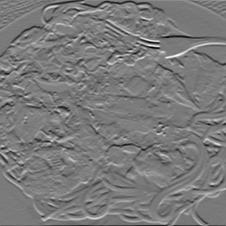}
\end{tabular}
\caption{Visualization of two images preprocessed with Sobel filter.
Sobel gives a $2$ channels output which at each point contain the vertical and horizontal derivative approximations.
Photographer usernames of these two YFCC100M RGB images are respectively \textit{booledozer} and \textit{nathalie.cone}.
}
\label{fig:preprocess}
\end{figure}

First, in Figure~\ref{fig:color_std}, we randomly select a subset of $3000$ clusters and sort them by standard deviation to their mean color.
If the standard deviation of a cluster to its mean color is low, it means that the images of this cluster tend to have a similar colorization.
Moreover, we show in Figure~\ref{fig:color_cluster} some clusters with a low standard deviation to the mean color.
We observe in Figure~\ref{fig:color_std} that the clustering on features learned with our method focuses more on color than the clustering on RotNet features.
Indeed, clustering by color and low-level information produces balanced clusters that can easily be predicted by a convnet.
Clustering by color is a solution to our formulation.
However, as we want to avoid an uninformative clustering essentially based on colors, we remove some part of the input information by feeding the network with the image gradients instead of the raw RGB image (see Figure~\ref{fig:preprocess}).
This allows to greatly improve the performance of our features when evaluated on downstream tasks as it can be seen in Table~\ref{tab:rgb}.
We observe that Sobel filter improves slightly RotNet features as well.

\begin{table}[t]
\centering
  \setlength{\tabcolsep}{3.5pt}
    \begin{tabular}{@{}l c c c@{}}
      \toprule
      Method & Data & RGB & Sobel \\
      \midrule
      RotNet      & YFCC 1M & $69.8$ & $70.4$ \\
      \midrule
      \OURS & YFCC 20M & $71.6$ & $76.1$ \\
      \bottomrule
    \end{tabular}
    \caption{
Influence of applying Sobel filter or using raw RGB input on the features quality.
We report validation mAP on Pascal VOC classification task (\textsc{fc68} setting).
    }
    \label{tab:rgb}
\end{table}

\section{Hyperparameters}
In this section, we detail our different hyperparameter choices.
Images are rescaled to $3 \times 224 \times 224$.
Note that for each network we choose the best performing hyperparameters by evaluating on Pascal VOC $2007$ classification task without finetuning.
\begin{itemize}
\item \textbf{RotNet YFCC100M}: we train with a total batch-size of $512$, a learning rate of $0.05$, weight decay of $0.00001$ and dropout of $0.3$.
\item \textbf{RotNet ImageNet}: we train with a total batch-size of $512$, a learning rate of $0.05$, weight decay of $0.00001$ and dropout of $0.3$.
\item \textbf{DeepCluster YFCC100M 1.3M images}: we train with a total batch-size of $256$, a learning rate of $0.05$, weight decay of $0.00001$ and dropout of $0.5$.
A sobel filter is used in preprocessing step. We cluster the pca-reduced to $256$ dimensions, whitened and normalized features with $k$-means into $10.000$ clusters every $2$ epochs of training.
\item \textbf{DeeperCluster YFCC100M}: we train with a total batch-size of $3072$, a learning rate of $0.1$, weight decay of $0.00001$ and dropout of $0.5$.
A sobel filter is used in preprocessing step. 
We cluster the whitened and normalized features (of dimension $4096$) of the non-rotated images with hierarchical $k$-means into $320.000$ clusters ($4$ clusterings in $80.000$ clusters each) every $3$ epochs of training.
\item \textbf{DeeperCluster ImageNet}: we train with a total batch-size of $748$, a learning rate of $0.1$, weight decay of $0.00001$ and dropout of $0.5$.
A sobel filter is used in preprocessing step. We cluster the whitened and normalized features (of dimension $4096$) of the non-rotated images with $k$-means into $10.000$ clusters every $5$ epochs of training.
\end{itemize}
For all methods, we use stochastic gradient descent with a momentum of $0.9$.
We stop training as soon as performance on Pascal VOC $2007$ classification task saturates.
We use PyTorch version 1.0 for all our experiments.

\section{Usernames of cluster visualization images}
For copyright reason, we give here the Flickr user names of the images from Figure~$5$.
For each cluster, the user names are listed from left to right and from top to bottom.
Photographers of images in cluster \textit{cat} are sun\_summer, savasavasava, windy\_sydney, ironsalchicha, Chiang Kai Yen, habigu, Crackers93, rikkis\_refuge and rabidgamer.
Photographers of images in custer \textit{elephantparadelondon} are Karen Roe, asw909, Matt From London, jorgeleria, Loz Flowers, Loz Flowers, Deck Accessory, Maxwell Hamilton and Melinda 26 Cristiano.
Photographers of images in custer \textit{always} are troutproject, elandru, vlauria, Raymond Yee, tsupo543, masatsu, robotson, edgoubert and troutproject.
Photographers of images in custer \textit{CanoScan} are what-i-found, what-i-found, allthepreciousthings, carbonated, what-i-found, what-i-found, what-i-found, what-i-found and what-i-found.
Photographers of images in custer \textit{GPS: (43, 10)} are bloke, garysoccer1, macpalm, M A T T E O 1 2 3, coder11, Johan.dk, chrissmallwood, markomni and xiquinhosilva.
Photographers of images in custer \textit{GPS: (-34, -151)} are asamiToku, Scott R Frost, BeauGiles, MEADEN, chaitanyakuber, mathias Straumann, jeroenvanlieshout, jamespia and Bastard Sheep.
Photographers of images in custer \textit{GPS(64, -20)} are arrygj, Bsivad, Powys Walker, Maria Grazia Dal Pra27, Sterling College, roundedbygravity, johnmcga, MuddyRavine and El coleccionista de instantes.
Photographers of images in custer \textit{GPS: (43, -104)} are dodds, eric.terry.kc, Lodahln, wmamurphy, purza7, jfhatesmustard, Marcel B., Silly America and Liralen Li.



\end{document}